\documentclass[sn-basic, Numbered]{sn-jnl}

\usepackage{graphicx}%
\usepackage{multirow}%
\usepackage{amsmath,amssymb,amsfonts}%
\usepackage{amsthm}%
\usepackage{mathrsfs}%
\usepackage[title]{appendix}%
\usepackage{xcolor}%
\usepackage{textcomp}%
\usepackage{manyfoot}%
\usepackage{booktabs}%
\usepackage{algorithm}%
\usepackage{algorithmicx}%
\usepackage{algpseudocode}%
\usepackage{listings}%
\usepackage[inline]{enumitem}
\usepackage{multirow}
\usepackage{float}
\usepackage{multicol}

\usepackage[nolist,nohyperlinks]{acronym} 
\usepackage{svg}
\usepackage{caption}
\usepackage{subcaption}
\usepackage{wrapfig}
\usepackage{soul,color}
\soulregister\cite7
\soulregister\ref7
\soulregister\pageref7

\usepackage[textwidth=25mm,textsize=footnotesize,disable]{todonotes}
\makeatletter
\if@todonotes@disabled

\else

\fi

\begin{document}

\title[Scalability of Reinforcement Learning Methods for Dispatching in Semiconductor Frontend Fabs: A Comparison of Open-Source Models with Real Industry Datasets]{Scalability of Reinforcement Learning Methods for Dispatching in Semiconductor Frontend Fabs: A Comparison of Open-Source Models with Real Industry Datasets}


\author*[1,2]{\fnm{Patrick} \sur{Stöckermann}}\email{patrick.stoeckermann@infineon.com}

\author[1,3]{\fnm{Henning} \sur{Südfeld}}

\author[1,4]{\fnm{Alessandro} \sur{Immordino}}

\author[1]{\fnm{Thomas} \sur{Altenmüller}}

\author[3]{\fnm{Marc} \sur{Wegmann}}

\author[2]{\fnm{Martin} \sur{Gebser}}

\author[2]{\fnm{Konstantin} \sur{Schekotihin}}

\author[5]{\fnm{Georg} \sur{Seidel}}

\author[6]
{\fnm{Chew Wye} \sur{Chan}}

\author[6]
{\fnm{Fei Fei} \sur{Zhang}}

\affil[1]{\orgname{Infineon Technologies AG}, \orgaddress{\street{Am Campeon 1-15}, \city{Neubiberg}, \postcode{85579}, \country{Germany}}}

\affil[2]{\orgdiv{Department of Artificial Intelligence and Cybersecurity}, \orgname{University of Klagenfurt}, \orgaddress{\street{Universitätsstraße 65-67}, \city{Klagenfurt am Wörthersee}, \postcode{9020}, \country{Austria}}}

\affil[3]{\orgdiv{Institute for Machine Tools and Industrial Management}, \orgname{Technical University Munich}, \orgaddress{\street{Boltzmannstr. 15}, \city{Garching b. München}, \postcode{85748}, \country{Germany}}}

\affil[4]{\orgdiv{Department of Information Engineering}, \orgname{University of Padua}, \orgaddress{\street{Via Gradenigo 6/B}, \city{Padova}, \postcode{35131}, \country{Italy}}}

\affil[5]{\orgname{Infineon Technologies Austria}, \orgaddress{\street{Siemensstraße 2}, \city{Villach}, \postcode{9500}, \country{Austria}}}

\affil[6]
{\orgname{D-SIMLAB Technologies Pte Ltd}, \orgaddress{\street{8 Jurong Town Hall Road, 23-05}, \city{Singapore}, \postcode{609434}, \country{Singapore}}}

\abstract{Benchmark datasets are crucial for evaluating approaches to scheduling or dispatching in the semiconductor industry during the development and deployment phases. However, commonly used benchmark datasets like the Minifab or SMT2020 lack the complex details and constraints found in real-world scenarios. To mitigate this shortcoming, we compare open-source simulation models with a real industry dataset to evaluate how optimization methods scale with different levels of complexity. Specifically, we focus on Reinforcement Learning methods, performing optimization based on policy-gradient and Evolution Strategies. Our research provides insights into the effectiveness of these optimization methods and their applicability to realistic semiconductor frontend fab simulations. We show that our proposed Evolution Strategies-based method scales much better than a comparable policy-gradient-based approach. Moreover, we identify the selection and combination of relevant bottleneck tools to control by the agent as crucial for an efficient optimization. For the generalization across different loading scenarios and stochastic tool failure patterns, we achieve advantages when utilizing a diverse training dataset. While the overall approach is computationally expensive, it manages to scale well with the number of CPU cores used for training. For the real industry dataset, we achieve an improvement of up to 4\,\% regarding tardiness and up to 1\,\% regarding throughput. For the less complex open-source models Minifab and SMT2020, we observe double-digit percentage improvement in tardiness and single digit percentage improvement in throughput by use of Evolution Strategies.
}


\maketitle

%
\begin{acronym}[SFSP]
\setlength{\itemsep}{-\parsep}

\acro{ABC}{Artificial Bee Colony}
\acro{ACK}{ACKnowledgement}
\acro{A3C}{Advantage Actor-Critic}
\acro{AI}{Artificial Intelligence}
\acro{API}{Application Programming Interface}
\acro{ASCII}{American Standard Code for Information Interchange }
\acro{CEO}{Chief Executive Officer}
\acro{CP}{Constraint Programming}
\acro{CT}{Cycle Time}
\acro{DABC}{Discrete Artificial Bee Colony Algorithm}
\acro{DIN}{Deutsches Institut für Normung}
\acro{DDPG}{Deep Deterministic Policy Gradient}
\acro{DLL}{Dynamic Link Library}
\acro{DDQN}{Deep Q-Network}
\acro{DQN}{Double Deep Q-Network}
\acro{EDD}{Eearliest Due Date}
\acro{EG}{Equipment Group}
\acro{GPU}{Graphics Processing Unit}
\acro{IEEE}{Institute of Electrical and Electronics Engineers}
\acro{IMPALA}{Importance Weighted Actor-Learner Architecture}
\acro{ISO}{International Organization for Standardization}
\acro{JSON}{JavaScript Object Notation}
\acro{LCS}{learning-based Cuckoo Search} 
\acro{NPN}{Schichtenabfolge: n-Schicht, p-Schicht, n-Schicht}
\acro{PNP}{Schichtenabfolge: p-Schicht, n-Schicht, p-Schicht}
\acro{RL}{Reinforcement Learning}
\acro{SFSP}{Semiconductor Factory Scheduling Problem}
\acro{JSSP}{Job-Shop Scheduling Problem}
\acro{GNN}{Graph Neural Network}
\acro{GTN}{Graph Transformer Network}
\acro{GWO}{Grey Wolf Optimizer}
\acro{PPO}{Proximal Policy Optimization}
\acro{DQN}{Deep Q-Network}
\acro{MIP}{Mixed Integer Programming}
\acro{MDP}{Markov Decision Process}
\acro{NN}{Neural Network}
\acro{ES}{Evolution Strategies}
\acro{CMA}{Covariance Matrix Adaptation}
\acro{CMA-ES}{Covariance Matrix Adaptation Evolution Strategies}
\acro{KPI}{Key Performance Indicator}
\acro{NES}{Natural Evolution Strategies}
\acro{LSF}{Load Sharing Facility}
\acro{DRL}{Deep Reinforcement Learning}
\acro{WIP}{Work In Progress}
\acro{FF}{Flow Factor}
\acro{SARSA}{State–Action–Reward–State–Action}
\acro{SRPT}{Shortest Remaining Processing Time}
\acro{SPT}{Shortest Processing Time}
\acro{SCFM}{Semiconductor Frontend Manufacturing}
\acro{SWOT}{Strengths, Weaknesses, Opportunities and Threats}
\acro{TRPO}{Trust Region Policy Optimization}
\acro{SMSP}{Single Machine Scheduling Problem}
\acro{PMSP}{Parallel Machine Scheduling Problem}
\acro{FSSP}{Flow Shop Scheduling Problem}
\acro{FFSSP}{Flexible Flow Shop Scheduling Problem}
\acro{JSSP}{Job Shop Scheduling Problem}
\acro{FJSSP}{Flexible Job Shop Scheduling Problem}
\acro{CJSSP}{Complex Job Shop Scheduling Problem}
\acro{SFSP}{Semiconductor Factory Scheduling Problem}
\acro{SFTSP}{Semiconductor Final Testing Scheduling Problem}
\acro{FIFO}{First-In First-Out }
\acro{CR}{Critical Ratio}
\end{acronym}

\section{Introduction}\label{sec1}

Scheduling and dispatching problems in semiconductor frontend manufacturing present significant challenges due to their complexity and optimization potential. However, most approaches lack comparability as they are evaluated on vastly different datasets. Even for the same dataset and approach, researchers can obtain varying results due to the different simulator engines used and the stochastic nature of these models, which can depend on random seeds. Therefore, we aim to compare multiple datasets and test two different \ac{RL} approaches on all of the datasets to evaluate scalability and optimization potential. As public benchmark scenarios often neglect important details of real manufacturing facilities, we additionally include a large-scale industry dataset in our tests.

With increasing complexity or problem size, \ac{RL} training is typically distributed by computing trajectories in many parallel environments. Besides classic policy-gradient \ac{RL} approaches like \ac{PPO}, \ac{ES} have been found to be a scalable alternative \cite{Salimans.2017}. Scalability of \ac{RL} systems also includes implementation effort \cite{Liang.2021}, where RLlib \cite{Liang.2018} offers an efficient solution to realize distributed training techniques for \ac{RL} systems, including automatic hyperparameter search.

In this paper, we compare different testbeds from the literature with real large-scale semiconductor manufacturing facilities. We then evaluate the potential and scalability of different \ac{DRL} approaches for dispatching lots in these scenarios.

\pagebreak

Our contributions can be summarized as follows:
\begin{itemize}
    \item We present the \textbf{first application of \ac{RL}} to \textbf{multiple public benchmark testbeds} as well as a \textbf{large-scale industrial-grade scenario} for lot dispatching in semiconductor manufacturing.
    \item We compare the \textbf{scalability} of RL algorithms \textbf{across different testbeds} and for \textbf{various tools} within individual scenarios.
    \item We analyze the \textbf{generalization properties} of the trained RL agents across different scenarios.
    \item We experimentally investigate the \textbf{computational cost} associated with the benchmark scenarios.
\end{itemize}

\section{Fundamentals and Related Work}\label{sec2}
    In order to better understand the different classes of scheduling problems in research and industry, we give an overview regarding different classifications for those problems and then explain the \acl{SFSP}. Furthermore, we explain the main concepts of \ac{RL} and introduce the two algorithms most relevant in this publication. 

    \subsection{Scheduling in Semiconductor Manufacturing}\label{sec:ppc}
        Scheduling problems are characterized by multiple constraints of different types (e.g., process technology or logistics), which affect the assignment of jobs to machines \cite{Klemmt.2012}. The most common scheduling problems will be introduced in the following.

        \textbf{\acp{SMSP}} are relevant for systems with an individual bottleneck that significantly impacts the overall system performance. Often the job sequence is optimized with static dispatching rules such as \ac{FIFO} or \ac{EDD} \cite{Pinedo.2005}. 

        The \textbf{\ac{PMSP}} deals with a number of parallel machines. This occurs in production environments where several stages or workcenters are arranged, each with multiple machines in parallel. The machines at a workcenter may be identical and a job can be processed on any of them \cite{Pinedo.2005}.

        \textbf{\acp{FSSP}} are typical in setups where multiple operations need to be performed on different machines. Each job has the same sequence of machines to go through. The \textbf{\ac{FFSSP}} is an extension of this problem with multiple parallel machines at each stage \cite{Pinedo.2005}.

        The \textbf{\ac{JSSP}} additionally has jobs with different orders regarding the sequence of operations. In the most simple version, each of $n$ jobs has to be processed once on each of $m$ machines \cite{Pinedo.2005}.

        In semiconductor manufacturing, the \textbf{\ac{FJSSP}} is typical and features workcenters with multiple machines in parallel \cite{Pinedo.2005}. 
        
        In case some jobs can be processed together at the same time on the same machine, it is a \textbf{batching problem}. Grouping jobs together to build batches becomes part of the scheduling problem in this case \cite{Brucker.2007}. 
        The \textbf{\ac{CJSSP}} incorporates, among other things, batching and reentrant flows. In a reentrant flow shop, the same machine can occur multiple times in the same route. This significantly increases complexity \cite{Knopp2017}.

        The properties of the described scheduling problems are compared in Table \ref{tab:scheduling_problems}.
        
        \textbf{\ac{SCFM}} includes hundreds of process steps, on hundreds of machines with highly diverse and stochastic processing times. It involves the coordination of $n$ diverse jobs and $m$ sophisticated machines in a non-preemptive queuing environment. Operations can be processed on multiple candidate machines constrained by setup requirements. Based on these traits, the \textbf{\ac{SFSP}} is often considered a complex job shop problem, with some models considering additional details. It is often decomposed into \ac{SMSP}, \ac{PMSP}, or \ac{FJSSP} in order to simplify implementation and reduce the computation time. Typically, only bottleneck machines, often the lithography area, are scheduled with traditional approaches using \ac{CP} or \ac{MIP} \cite{Mason.2022, Moench.2011}. 

        \begin{table}[t]
            \caption{Comparison of scheduling problems}
            \label{tab:scheduling_problems}
            \begin{tabular}{lccccccc}
            \hline
                                    & SMSP & PMSP & FSSP & FFSSP & JSSP & FJSSP & CJSSP \\
            \hline
            parallel machines       &      & x    &      & x     & x    & x     & x     \\
            sequential   operations &      &      & x    & x     & x    & x     & x     \\
            different   sequences   &      &      &      &       & x    & x     & x     \\
            batching                &      &      &      &       &      & x     & x     \\
            reentrant flow          &      &      &      &       &      &       & x    \\
            \hline
            \end{tabular}
        \end{table}

        The ($\alpha \| \beta \| \gamma$) notation introduced by \cite{Graham.1979} is even more elaborate and defines $\alpha$ as the type of machine environment defined by one of the scheduling problem definitions introduced above. Information about the jobs and sequencing constraints is given by $\beta$. The objective function is given by $\gamma$. More details can be found in \cite{Klemmt.2012}.

        We can further differentiate between deterministic and stochastic scheduling. In deterministic scheduling, processing times, setup times, transport times, due dates and other parameters are assumed to be known in advance and not influenced by uncertainty. However, there are many sources of uncertainty in a real semiconductor manufacturing environment. One of the biggest sources of uncertainty are machine breakdowns. In stochastic scheduling, probabilistic values are not assumed to be known in advance. They are replaced by probability distributions modeling the variability of reality. \cite{Monch.2013,Pinedo.2012} 
        
        Even the classic \ac{JSSP} is NP-hard \cite{Applegate1991}, and all more advanced variants as well \cite{Gupta2004}.  In order to understand the scalability of deterministic methods, \cite{DaCol.2022} analyzed the performance of different \ac{CP} solvers on scenarios with up to a million operations to be scheduled on up to one thousand machines. However, they did not consider the constraints that are required in an actual industry use case such as the dedications of processes to a subset of machines within a tool group, batch processes with complex recipe-related constraints and a much more diverse load-mix. Even if the algorithm would be capable of computing a solution within a few hours for a realistic dataset, the solution is already too old after some minutes, as the situation in a real facility constantly changes due to varying processing and transport times as well as failed processes and machine breakdowns. Often, manufacturers use inflexible and suboptimal handcrafted heuristics for dispatching to circumvent the problems of mathematical scheduling.
    
    \subsection{Semiconductor Frontened Manufacturing Testbeds}  
        In order to evaluate deterministic scheduling approaches without any stochastic modelling \cite{DaCol.2022, Tassel.2021}, researchers often use \ac{JSSP} benchmark suites like Taillard's \cite{Taillard.1993} and solvers such as the open-source software OR-Tools \cite{Perron.2023} or the commercial solution IBM CPLEX \cite{Laborie.2018}.
   
\begin{figure}[t]
            \centering
            \includegraphics[width=0.99\textwidth]{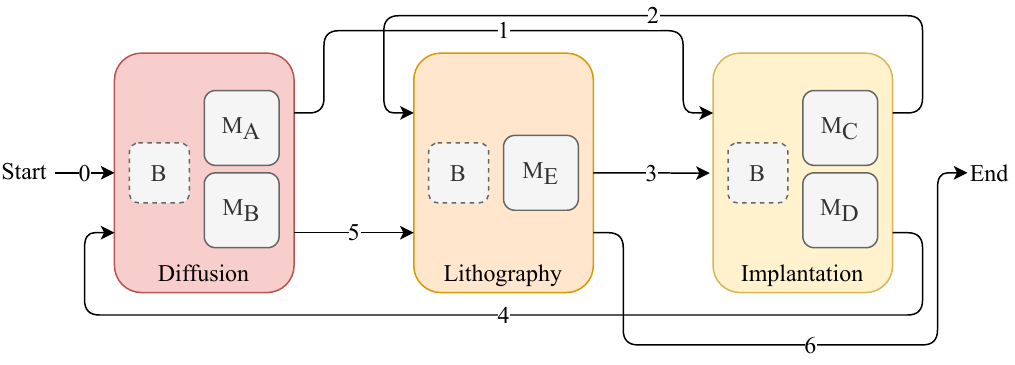}
            \caption{The Minifab model layout, adapted from \cite{ElAdl.1996}.}
            \label{fig:Minifab}
        \end{figure}

        However, we are focusing on stochastic models that are much closer to the properties found in real semiconductor manufacturing facilities.
        One of the first and widely used testbeds for \ac{SCFM} scheduling problems is the Minifab model which was developed by researchers from Intel and Arizona State University in an attempt to bridge the gap between research and industry. It incorporates routes with 6 steps on 5 machines. More specifically, it entails two implantation, one lithography and two diffusion tools. The latter are additionally capable of batching three lots at a time together. The fab incorporates three different products with different release rates. Generally, the model aims to provide a variant of the \ac{CJSSP} with stochastic influences such as machine breakdowns and preventive maintenance. The Minifab model layout \cite{ElKhouly.2009, Spier.1995} is shown in Figure \ref{fig:Minifab}. 

        The Semiconductor Manufacturing Testbed 2020 (SMT2020) is an extension of the Measurement and Improvement of Manufacturing Capacity (MIMAC) datasets \cite{Fowler.1995}, which was developed by SEMATECH in 1995. SMT2020 incorporates a larger factory scale with more tools and processing steps. Furthermore, preventive maintenance and unscheduled downtime are modelled. The authors of SMT2020 also note that the older MIMAC datasets miss clear implementation guidelines \cite{Kopp.2020}, and therefore it is hard to reproduce results.
        
        The SMT2020 fab models include scenarios for custom orders (Low-Volume/High-Mix) and mass production (High-Volume/Low-Mix). More than 500 operations are modelled for the individual and diverse product routes, performed by between 1071 and 1265 machines depending on the scenario. The transport times are accounted for and operation- as well as sequence-dependent setups are modelled respecting tool specific requirements of the semiconductor manufacturing process. 
        Moreover, the SMT2020 fab models include batching policies for some operations and probabilistic skipping of quality assessments. Critical jobs, called ``Hot Lots'' and ``Super-Hot Lots'', are prioritized to meet urgent deadlines. \ac{SFSP} adapts to real-time issues, such as defect detection leading to job rerouting \cite{Kopp.2020, Tassel.2023}.

        Additional models worth mentioning are the recently published Complex Job Shop Simulation reference model (CoJoSim) \cite{Bauer.2023}, which offers a reference implementation based on the MIMAC datasets as well as the smaller Sematech 300mm model \cite{Campbell.2000} and the scaled down model of a Harris Corporation fab \cite{Kayton.1997}.  
        
        \begin{table}[t]
        \caption{Comparison of reference models adapted from \cite{Bauer.2023}.}\label{tab:model_comparison}%
        \begin{tabular}{llllll}
        \hline
        \multicolumn{1}{c}{Model} & MIMAC     & Minifab & Harris   & SEMATECH 300mm & SMT2020  \\ \hline
        No. machines              & up to 260 & 5       & 12       & 275            & 1043       \\
        No. machine groups        & up to 85  & 3       & 11       & 103            & 105      \\
         No. products              & up to 21  & 2       & 3        & 1              & 10        \\
        No. process steps         & up to 280 & 6       & up to 22 & 364            & up to 632 \\
        \hline
        \end{tabular}
        \end{table}
        The mentioned models are compared to each other in Table \ref{tab:model_comparison}.
        In SMT2020, the scale of the addressed problem goes well beyond the smaller test datasets. 
        However, the SMT2020 fab models have some shortcomings compared to real manufacturing data. We therefore introduce a real industry scenario of similar size but with a more diverse load mix and complex tool dedications. Similar to SMT2020, our dataset has more than 1000 machines, but contains more than ten times the number of products considered in SMT2020. Another important detail that is missing in SMT2020 is that a specific operation may not be executable on all tools within a group and can have flexible processing times on different tools. Especially the dedication of processes to a subset of machines within a tool group in combination with a more diverse load mix leads to a sharp increase in constraints. This also affects batch processes with complex recipe-related constraints limiting the type of products which can be batched together. 
        In fact, the homogeneity between tools of the same group assumed by SMT2020 drastically reduces the complexity. For AI-based methods, extracting statistical relationships and patterns in the data becomes much more difficult with a more diverse dataset. Furthermore, we consider multiple heterogeneous loading scenarios with varying load-mix and volume for the industry case. This significantly increases the difficulty of generalization, as the dispatching policy must not overfit on the training scenarios. Furthermore, the increased complexity leads to longer execution time for the simulation, increasing the computational cost of training.\\
        Lastly, our real industry scenario has equipment group specific combinations of dispatching heuristics, fine tuned by domain experts. These rules are the most difficult to beat among the different models. \cite{Stockermann.2023}
        
    \subsection{Reinforcement Learning}\label{subsec2}
        \ac{RL} enables the autonomous learning of optimized decision-making policies to act in complex environments. This is possible through continuous learning by interacting with an environment, i.e., a simulation.
        Usually, the \ac{MDP} is used to formalize \ac{RL} environments \cite{Sutton.1998}. \ac{RL} algorithms typically optimize a cumulative reward $R$ over a number of steps $T$ instead of optimizing for individual direct returns $r_t$ for each time step $t$:
            \begin{equation*}
                R=\sum_{t=0}^{T-1} \gamma^t \cdot r_{t+1}
            \end{equation*}
        The individual rewards are weighted by introducing a discount factor $\gamma$, which determines the importance of immediate and future rewards.
        There are numerous \ac{RL} algorithms with different advantages and weaknesses. Some of the most popular derivative-based \ac{RL} algorithms, such as policy-based \ac{PPO} \cite{Schulman.2017, Mnih.2016} or value-based \ac{DQN} \cite{Mnih.2013} approaches, have also been used for dispatching and scheduling in semiconductor manufacturing \cite{Tassel.2021, Altenmueller.2020}.
        Furthermore, taking \ac{ES} \cite{Salimans.2017} as an alternative to derivative-based \ac{RL} algorithms also attracted interest in scheduling applications \cite{Tassel.2023, Stockermann.2023}.

        \ac{PPO} \cite{Schulman.2017} is based on the \ac{TRPO} \cite{Schulmann.2015} algorithm and significantly simplifies implementation. The stochastic policy is defined as $\pi_\theta$, where $\theta \in \mathbb{R}^n $ is a vector of parameters. The policy $\pi_{\theta}(a_t \mid s_t)$ defines the probability to take an action $a_t$ in the state $s_t$ at time step $t$. The 
        likelihood ratio $\rho_t(\theta)=\frac{\pi_{\theta}\left(a_t \mid s_t\right)}{\pi_{\theta_{\text {old }}}\left(a_t \mid s_t\right)}$, with $\theta_{old}$ denoting the policy parameters, is used during sampling. This allows to define the PPO objective, which is called clipped surrogate objective, as
        $$
        \mathcal{L}^{\mathrm{CLIP}}(\theta)=\hat{\mathbb{E}}_t\left[\min \left(\rho_t(\theta) \hat{A}_t, \operatorname{clip}\left(\rho_t(\theta), 1-\epsilon, 1+\epsilon\right) \hat{A}_t\right)\right]
        $$
        with the hyperparameter $\epsilon \geq 0$ and the advantage estimation $\hat{A}_t$ for time step $t$. The approach ensures that \ac{PPO} does not perform too radical updates to the parameters of the policy. In its Actor-Critic variant, the parameter vector $\phi \in \mathbb{R}^m$ defines the \ac{NN} of the state-value function estimator, also called critic, and $\theta$ represents the parameters of the policy network, also called actor. A simplified version of our \ac{PPO} implementation, which is based on Ray RLlib \cite{Liang.2018}, is shown in Algorithm~\ref{alg:ppo} with $N$ parallel agents and data buffer $\mathcal{D}$.

        \begin{algorithm}[t]
            \caption{PPO}
            \label{alg:ppo}
            \begin{algorithmic}[1]
                \State Initialize $\theta_{\text {old }}, \phi_{\text {old }}$
                \For{iteration\texttt{ $t=1,2,\ldots$ }}
                    \State Synchronize networks parameters among all the $N$ agents
                    \State $\mathcal{D}=\emptyset$ 
                    \For{agent \texttt{$t=1,\ldots,N$ }}
                        \State Collect a trajectory $\tau$ of length $T$ using $\pi_{\theta_{\text {old }}}$
                        \State Post-processs $\tau$ to get the advantage estimation for each sample
                        \State $\mathcal{D}=\mathcal{D} \cup \tau$
                    \EndFor
                    \State Optimize $\mathcal{L}$ w.r.t. $\theta, \phi$ by performing $K$ epochs with mini-batch size $M$ over $\mathcal{D}$
                    \State $\theta_{\text {old }} \leftarrow \theta$
                    \State $\phi_{\text {old }} \leftarrow \phi$
                \EndFor
            \end{algorithmic}
        \end{algorithm}
        
        \ac{ES} can be seen as a black-box algorithm and thus solves the credit assignment problem \cite{Sutton.1998} typical for most \ac{RL} approaches. The agent only receives one reward for each episode, which is well-suited for environments where a good value function estimate is hard to derive and the effects of individual actions apply with substantial delay. The algorithms add Gaussian noise, with the mean $\mu$ and fixed covariance $\sigma^2I$ in the most simple variant, to the parameters instead of using back-propagation. This leads to good exploration, easy implementation, but poor scalability with the number of parameters~$\theta$.

        We make use of the \ac{CMA} approach \cite{Hansen.2001} utilized by \cite{Stockermann.2023}, which promises faster convergence by using the covariance of the parameters~$\theta$ for sampling.
        Algorithm \ref{alg:cames} outlines the \ac{CMA}-\ac{ES} approach, where $\alpha_\mu$ controls the rate at which the mean $\mu$ is updated. $N_{best}$ is a hyperparameter that determines the subset of the most promising candidates $w_i$ based on their score given by the fitness function $F(w_i)$. At each iteration $t$, multiple candidate solutions created by sampling are tested in parallel. The returns are then used to approximate the fitness function and guide the sampling for the next iteration.
        \begin{algorithm}[t]
            \caption{Summarized method for \ac{CMA}-\ac{ES} presented by \protect\cite{Hansen.2001}}
            \label{alg:cames}
            \begin{algorithmic}[1]
                \State \textbf{Input} : Step size $\sigma_{0}$, covariance matrix $C_{0} = I$, mean vector $\mu_{0}$, number $N_{best}$ of most promising solutions considered for optimization, number $n$ of parallel agents, maximum iterations $T$
                \For{\texttt{$t=1,\ldots,T$ }}
                    \State Sample $w_{1},...w_{n}\sim N(\mu_{t},\sigma^2 C_{t})$
                    \State Compute return values $F_{i} = F(w_{i})$ for $i=1, \ldots, n$
                    \State $\sigma_{t+1} \leftarrow \frac{1}{N_{best}} \sum^{N_{best}}_{i=1} (w_{i} - \mu_{t})$
                    \State $\mu_{t+1}\leftarrow\frac{1}{N_{best}}\sum^{N_{best}}_{i=1}  w_{i}$
                \EndFor
            \end{algorithmic}
        \end{algorithm}
    
    \subsection{Reinforcement Learning in Semiconductor Frontened Manufacturing}\label{subsec3}
        We conduct a comprehensive structured literature search in order to find relevant publications in the field of RL-based dispatching or scheduling approaches in semiconductor manufacturing based on \cite{Webster.2022}. We execute the search using the abstract and citation database Scopus \cite{Scopus.2025} with the following search string:
        \begin{lstlisting}[xleftmargin=\dimexpr-\leftmarginii-\leftmargini,basicstyle=\sf]
        ( TITLE ( ``Dispatch*'' OR ``Schedul*'' ) OR ABS ( ``Dispatch*'' OR ``Schedul*'' ) ) AND ( TITLE ( ``Semiconductor Frontend Manufacturing'' OR ``Semiconductor Production'' OR ``Semiconductor Manufacturing'' OR ``Wafer fab*'' ) OR ABS ( ``Semiconductor Frontend Manufacturing'' OR ``Semiconductor Production'' OR ``Semiconductor Manufacturing'' OR ``Wafer fab*'' ) ) AND ( TITLE ( ``Reinforcement Learning'' OR ``RL'' ) OR ABS ( ``Reinforcement Learning'' OR ``RL'' ) )
        \end{lstlisting}
        
        Publications from 2023 and before (1 year ago) must have at least 1 citation and 5 before 2020 (5 years ago) to be included due to significance. Next, we filter out everything that is not related to lot dispatching in a system with more than one modeled equipment group. These filtering steps reduce the results to 23 in Table \ref{tab:optimization_comparison_new}. 
        
        \begin{table}[t]
            \caption{Comparison of \ac{RL}-based scheduling and dispatching approaches in the literature sorted by citations (descending). The maximum size is reported with the number $m$ of machines and the number of equipment groups (EGs), if applicable.}
            \label{tab:optimization_comparison_new}
            \begin{tabular}{llllll}
            \hline
            \multicolumn{1}{c}{Year} & Author                       & Model(s)         & Approach      & Max Size         \\ \hline
            2018   & Waschneck et al.\ \cite{Waschneck.2018b}       & Custom        & DQN \cite{Mnih.2013}          & $m=6$, 4 EGs      \\
            2020   & Park et al.\ \cite{Park.2020}                  & Custom        & DQN \cite{Mnih.2013}          & $m=175$    \\
            2018   & Stricker et al.\ \cite{Stricker.2018}          & Custom        & DQN \cite{Mnih.2013}          & $m=8$, 3 EGs     \\
            2018   & Waschneck et al.\ \cite{Waschneck.2018}        & Custom        & DQN \cite{Mnih.2013}          & $m=6$, 4 EGs      \\
            2020   & Altenmüller et al.\ \cite{Altenmueller.2020}   & Custom        & DQN \cite{Mnih.2013}           & $m=10$, 5 EGs      \\
            2021   & Chien et al.\ \cite{Chien.2021}                & Custom        & DQN \cite{Mnih.2013}          & $m=9$      \\
            2022   & Lin et al.\ \cite{Lin.2022}                    & Custom, Wu \cite{Wu.2012}   & GWO \cite{Mirjalili.2015}          & $m=100$ \\
            2023   & Lin et al.\ \cite{Lin.2023}                    & YFJS-series \cite{Birgin.2013}                & LCS \cite{Yang.2010}          & $m=26$ \\
            2023   & Sakr et al.\ \cite{Sakr.2023}                  & Custom        & DQN \cite{Mnih.2013}           & $m=?$, 23 EGs     \\
            2022   & Liu et al.\ \cite{Liu.2022}                    & MiniFab \cite{Spier.1995}                     & A3C \cite{Mnih.2016}     & $m=5$, 3 EGs        \\
            2020   & Shiue et al.\ \cite{Shiue.2020}                & SEMATECH$^1$ \cite{Campbell.2000}             & Q-Learning \cite{Sutton.1998}          & $m=?$, 12 EGs \\
            2019   & Lee et al.\ \cite{Lee.2019}                    & Custom        & SARSA \cite{Sutton.1998}       & $m=160$ \\
            2023   & Tassel et al.\ \cite{Tassel.2023}              & SMT2020 \cite{Kopp.2020}                      & ES \cite{Salimans.2017}        & $m=1265$, 106 EGs   \\
            2024   & Lu \cite{Lu.2024}                         & Custom        & DDQN \cite{Hasselt.2016}  & $m=50$   \\
            2018   & Wangl et al.\ \cite{Wangl.2018}           & Custom        & Custom           & $m=220$ \\ 
            2023   & Ma et al.\ \cite{Ma.2023}                      & MiniFab \cite{Spier.1995}     & DDQN \cite{Hasselt.2016}     & $m=5$, 3 EGS\\
            2023   & Stöckermann et al.\ \cite{Stockermann.2023}    & Custom        & CMA-ES \cite{Salimans.2017, Hansen.2001}       & $m>1000$    \\
            2023   & Liao et al.\ \cite{Liao.2023}                  & Custom$^2$, TA$^2$ \cite{Taillard.1993}    & PPO \cite{Schulman.2017}           & $m=20$ \\
            2024$^3$ & Yedidsion \cite{Yedidsion.2024}         & Custom$^2$        & PPO \cite{Schulman.2017}          & $m=4$, 3 EGs      \\
            2024$^3$ & Zhangg \cite{Zhang.2024}                & \acs{SFTSP}$^2$ \cite{Wu.2008}        & Q-Learning \cite{Sutton.1998}          & $m=36$      \\ 
            2024$^3$ & Dong \cite{Dong.2024}                    & TA$^2$ \cite{Taillard.1993}, DMU$^2$ \cite{Demirkol.1998}       & \acs{GTN} \cite{Yun.2020}+\acs{ABC} \cite{Karaboga.2005}         & $m=20$     \\
            2024$^3$ & Wang \cite{Wang.2024}                    & Custom       & DDQN \cite{Hasselt.2016}          & $m=18$      \\
            2024$^3$ & Shao \cite{Shao.2024}                    & Custom        & A2C \cite{Mnih.2016}          & $m=10$, 4 EGs      \\
            \hline
            \end{tabular}
            \footnotetext[1]{Simplified version}
            \footnotetext[2]{No stochasticity modeled}
            \footnotetext[3]{Recently published, no citations yet}
        \end{table}
        
        It becomes obvious that it is very hard to compare the approaches as they were tested on many different and often custom models. 
        A problem that all of the presented models have in common is the comparability of the implementation, even for the same model \cite{Bauer.2023}.
        Some researchers implement their simulator with Python packages like SimPy \cite{Scherfke.2017}, build a new simulator from scratch \cite{Kovacs.2022, Tassel.2023}, or rely on a commercial simulator like  Autosched AP \cite{Phillips.1998}, Siemens Tecnomatix Plant Simulation \cite{bangsow.2020, Bauer.2023}, or D-Simlab D-Simcon Forecaster \cite{dsimcon.2023, Stockermann.2023}.
        Even with the same implementation, individual runs can vary a lot due to the stochastic events that heavily depend on the random seed. Furthermore, the hardware and computational resources vary between research labs. We therefore aim to conduct comparable experiments across the Minifab model, SMT2020 and an industrial scenario, all using the D-Simlab D-Simcon Forecaster \cite{dsimcon.2023}.

\section{Approach and Design of Experiments}\label{sec3}
    In this section, we introduce our overall approach as well as the design of experiments. This includes the utilized architecture, the simulator, the \ac{NN}, and the most important parts of the RL system, namely the policy, observation, action, and reward concepts.
    \subsection{Simulator Architecture}
\begin{figure}[t]
        \centering
        \includegraphics[width=0.99\textwidth]{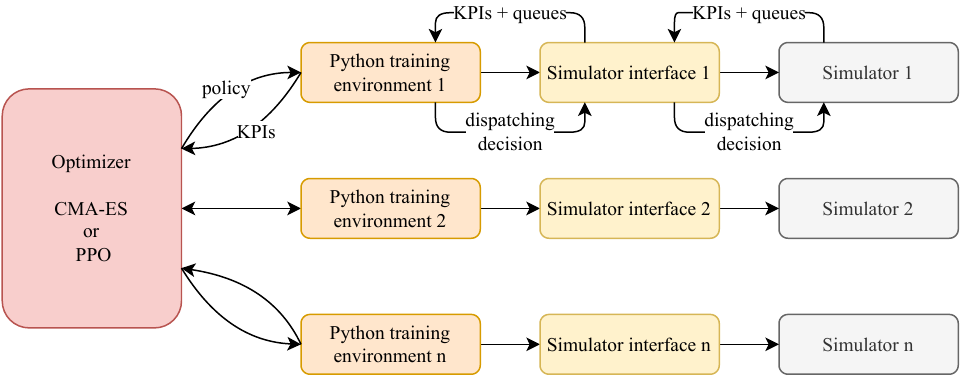}
        \caption{Architecture of experiments using the \ac{CMA}-\ac{ES} or \ac{PPO} optimizer, respectively.}
        \label{fig:optimizer_architecture}
    \end{figure}

     In order to conduct our experiments, we create a highly flexible training architecture with interchangeable simulation models and optimizers (see Figure~\ref{fig:optimizer_architecture}). The training is always controlled by the optimizer, which can be switched between the \ac{PPO} and CMA-ES algorithms described above.
     The optimizer then creates parallel workers running on separate CPUs.

     Each worker has a Python environment defining the current policy and thus the interaction with the simulator. The environment collects the observations of the state at every decision step to compute a dispatching decision, using the policy and providing the information needed for reward calculation. The individual environments are interacting with separate simulator instances through an interface. Each simulation can be individually initialized with different simulation models and different random seeds, defining stochastic events like tool failure.

     We use D-SIMLAB's highly realistic, state-of-the-art D-SIMCON simulator \cite{dsimcon.2023}, which can handle our industry scenario with over a thousand pieces of equipment as well as the open-source models Minifab and SMT2020. The simulator is calling the environment every time a dispatching decision has to be made and provides information about the simulator state, such as overall KPIs and queues of waiting lots.

     In order to calculate the tardiness of lots, we need due dates for each lot, which are the planned dates for completion. We define the tardiness as the time between the due dates and the real completion time. If a lot is behind schedule, the tardiness is the difference between these two dates, or zero in case the lot is ahead of schedule. The tardiness of uncompleted lots is calculated using their step due date, calculated from the final due date by subtracting the planned cycle time of remaining steps. The authors of SMT2020 already provide due dates in their dataset, and the same is the case for our industry dataset. For MiniFab, this detail is missing. Here we generate the due dates by multiplying the raw processing time with a planned \ac{FF} to obtain the planned \ac{CT} and thus also the due date. The planned FF is sampled from a uniform distribution between 2.1 and 2.5. We furthermore modify the processing times for each tool in the MiniFab model to vary between products. This condition makes the model more realistic.

     In the case of \ac{ES}, the policy of the individual environments is only updated once every episode, which is an entire simulation run simulating multiple weeks in the fab. In the case of \ac{PPO}, the optimizer is collecting frequent samples of state, action, and reward tuples to provide many yet small intermediate updates during an episode. \ac{PPO} therefore has much more synchronization and respective communication overhead.

    \subsection{Observation and Policy}
        We extend on the experiments of \cite{Stockermann.2023} by utilizing the \ac{CMA}-\ac{ES} optimizer \cite{Salimans.2017} for the optimization of Deep \acp{NN} in \ac{RL}. Additionally, we adapt the approach to use it with \ac{PPO}.

        Each simulator registers a predefined list of tools that are controlled by the agent at the beginning of the training.  The agent is called every time a dispatching decision has to be performed at one of those tools and receives representations of the lots that are waiting in front of the current tool. The entire queue of lots at the work center, to which the tool is assigned, is filtered to identify lots that can be processed by that specific tool. The queue of lots is represented using the following attributes, with the last two only used in case batch tools are controlled:
        \begin{multicols}{2}
            \begin{itemize}
                \item Time to fab due date
                \item Time to step due date
                \item Waiting time at current step
                \item Number of alternative tools
                \item Boolean flag indicating whether \\a faster tool exists
                \item Expected processing time
                \item Setup time
                \item Expected remaining cycle time 
                \item Number of wafers in lot
                \item Boolean flag indicating whether \\it is a batch tool
                \item Percentage of available wafers \\to form full batch
            \end{itemize}
        \end{multicols}
        All attributes are subject to $z$-score normalization using continuously updated statistics collected from all simulations during the experiment. 
        The policy is then used to compute a score for each lot. \ac{ES} picks the lot with the highest score and \ac{PPO} samples from a distribution, which is giving the probability of each lot based on the computed score. In case multiple lots can be dispatched at the same time, which is the case for batching tools, the algorithms selects lots that can be batched together with the lot with the highest score. Those additional lots are also selected based on their score. This is only done for \ac{ES}, though, as \ac{PPO} is based on individual discrete decisions and expects one index as the result of each forward pass. Batching tools are therefore not considered for the \ac{PPO} experiments given that the algorithm cannot handle such an implementation without major modifications.
        
        In order to be queue size-independent and capture the relationships between the lots in the queue, the observation is fed to the network lot by lot and combined as the dot product of projections. This is possible by the use of the Scaled Dot-Product Attention \cite{Vaswani.2017}. 
        It requires three projections for the queries, keys, and values and is computed by applying the softmax function on the scaled transposed dot product of the query $Q$ with all keys $K$. The transposed dot product is scaled by the dimension $d_k$ of the key projection, and the softmax result is multiplied with the value $V$:

        \begin{equation*}
        \operatorname{Attention}(Q, K, V)=\operatorname{softmax}\left(\frac{Q K^T}{\sqrt{d_k}}\right) V
        \end{equation*}

\begin{figure}[t]
            \centering
            \includegraphics[width=0.99\textwidth]{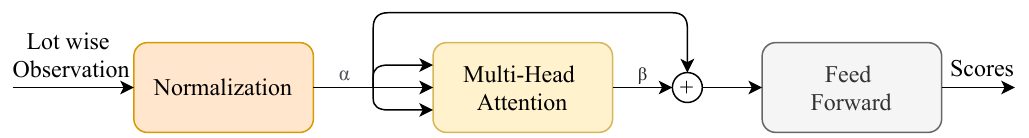}
            \caption{The architecture of the \ac{NN} for \ac{ES}.}
            \label{fig:attention_network}
        \end{figure}

        After the information of each lot is infused into the representation of the other lots, the modified representations are fed individually through a feed forward fully connected \ac{NN} in order to compute the score for each lot (see Figure~\ref{fig:attention_network}). The architecture is purposefully small in order to keep the number of trainable parameters low, as \ac{ES} does not scale well with the number of parameters.
        
\begin{figure}[b]
            \centering
            \includegraphics[width=0.99\textwidth]{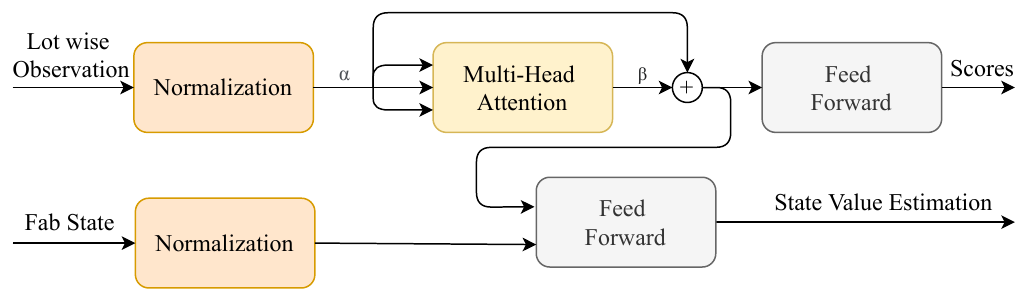}
            \caption{The architecture of the \ac{NN} for \ac{PPO}.}
            \label{fig:attention_network_ppo}
        \end{figure}

        For \ac{PPO}, we are additionally estimating the value function for the critic in order to compute the advantage for the \ac{PPO} loss function (see Figure~\ref{fig:attention_network_ppo}). This is important as our policy updates as well as the updates to the reward are much more frequent than for the evolutionary approach.
        This highlights the strength and weakness of PPO in one: if we are able to accurately estimate the value of the intermediate states during the episode, this should lead to more stable and faster convergence compared to updating the policy only once every episode. However, in case this estimate is noisy or incorrect, we cannot converge to a stable and valuable policy.
        
        \pagebreak
        
        The additional inputs representing the fab state are summarized as follows:
        \begin{multicols}{2}
            \begin{itemize}
                \item Work center WIP
                \item Completed wafers
                \item Total tardiness
                \item Total tardy lot count
                \item Average tardiness
                \item Standard deviation tardiness
                \item Average cycle time 
                \item Average fab WIP
            \end{itemize}
        \end{multicols}

        All values are compared to the respective values of the reference runs at the same time and given as the difference to the agent.

    \subsection{\ac{ES} Cost Function and \ac{PPO} Reward}   
        \label{sec:reward}
        We only assign one cost value $c_{\mathrm{ES}}$ per episode for the ES experiments. It is normalized by dividing the resulting KPIs of the training episode by the KPIs of the reference run using the default dispatching heuristics. As we mainly want to optimize the tardiness of lots, we base the cost on the tardiness $td_{out}$ of the completed wafers. In order to prevent the agent from sacrificing the other KPIs, we introduce conditional terms to the cost calculation that are only used if we get worse performance on the KPIs of the inner tardiness $td_{in}$ and the throughput $tp$. The throughput is the number of completed wafers for the entire episode. We set $\alpha_1 = \alpha_2 = 10$ in order to prevent the agent from accepting small worsening of $td_{in}$ and $tp$ to improve $td_{out}$:
        \begin{equation*}
            \label{eq:minifab_reward}
            \begin{aligned}
                c_{\mathrm{ES}}^* &=\frac{td_{out}}{td_{out,ref}} \\ 
                c_{\mathrm{ES}}&= 
                \begin{cases}
                    c_{\mathrm{ES}}^*\cdot \alpha_1 \frac{td_{in}}{td_{in,ref}}  & td_{in} > td_{in,ref} \\ 
                    c_{\mathrm{ES}}^*\cdot \alpha_2 \frac{tp_{ref}}{tp}  & tp < tp_{ref}
                \end{cases}
            \end{aligned}
        \end{equation*}

        For our industry scenario and its complex interdependencies, we utilize an adjusted cost function $c_{\mathrm{ES}}'$, as $c_{\mathrm{ES}}$ does not lead to satisfying results for that model. This seems to be due to the fact that is is way harder to maintain or even improve the throughput for the real-world scenario: 
        \begin{equation*}
            \begin{aligned}
            c_{\mathrm{ES}}' = \frac{td_{out}+td_{in}}{td_{out,ref}+td_{in,ref}} \cdot \left(\frac{tp_{ref}}{tp}\right)^2
            \end{aligned}
        \end{equation*}

        For \ac{PPO}, we assign a reward at each step based on the rolling mean of the same metrics as for the \ac{ES} approach. The rolling mean is recalculated every hour over the past 24 hours in order to reduce noise. Hence, the reward $r_{\mathrm{PPO},t}$ for time steps~$t$ is identical if the steps occur within the same hour: 
        \begin{equation*}
            r_{\mathrm{PPO},t} = \frac{tp_t}{(tp_t \cdot td_{out,t} + \mathrm{WIP}_t \cdot td_{in,t})\cdot (\mathrm{WIP}_t + tp_t)}
        \end{equation*}
        with $tp_t$ being the number of completed wafers during the past 24 hours and $\mathrm{WIP}_t$ the number of wafers remaining in the system at the end of the 24-hour interval. The reward is based on the division of the hourly throughput by the tardiness. The tardiness is the weighted sum of the tardiness of the wip and the tardiness of the wafers completed in the past 24\,hours.\\ As the \ac{PPO} approach receives regular updates of the reward, not only at the end of an episode, and should understand the relationship between individual actions and the reward, we design it continuously without case distinctions such as for $c_{\mathrm{ES}}$. The reward definition $r_{\mathrm{PPO},t}$ has been empirically found to be the best performing one in our experiments, additional rewards are compared in the appendix under Section \ref{sec:ppo_appendix}.
  
\section{Results}\label{sec4}
    In this section, we investigate the performance of simulation models with different static dispatching rules. Then, we select the best dispatching rule as a baseline and reference for comparing the \ac{RL} training techniques. Moreover, we analyze the impact of the utilized computational resources as well as the generalization capabilities.
    
    \subsection{Comparison of the Models' Default Dispatching Rules}

    For the Minifab model, we consider the following dispatching heuristics:
    \begin{itemize}
        \item \acf{FIFO}
        \item \acf{CR} (dividing remaining time to due date by the expected time to complete a lot)
        \item \acf{SRPT} (for the 
                          remaining steps)
        \item \acf{SPT} (for the next step)
        \item \acf{EDD}
    \end{itemize}
    Table~\ref{tab:heuristics_Minifab} shows (normalized) results
    regarding the number of completed wafers and the tardiness of lots. As the \ac{SRPT} heuristic dominates in terms of both performance metrics, we select it as baseline for the Minifab model.
    Here and for all following plots and tables, the tardiness $td$ is summed
    over all completed lots (CL), relative to their final due dates (DD), as well as
    uncompleted lots (WIP) relative to their step due dates (SDD):
    \begin{equation*}
            td = td_{in} + td_{out} = \sum_{l \in \mathrm{WIP}} \mathrm{SDD}_l - t + \sum_{l \in \mathrm{CL}} \mathrm{DD}_l - tc_l
        \end{equation*}
    considering the current time $t$, in this case the end time of the simulation, and the time of completion $tc$ for each finished lot $l$.
    Note that we want to increase the number of completed wafers, but decrease the tardiness.
    
    The authors of the SMT2020 fab models propose a hierarchical combination of dispatching heuristics \cite{Kopp.2020}, which is also typical in the industry.
    Lots that are currently under time constraints, i.e., they will have to repeat the previous operation if the next step is not performed in time, are always processed first. Prioritized lots are dispatched next: Super-Hot Lots at the highest priority, followed by Hot Lots. Third, jobs that align with the current machine setup are favored as this prevents unnecessary setup time from reducing the throughput. Finally, a tie-breaking heuristic is applied, where we consider the same five dispatching heuristics as also used for the Minifab model. 

    \begin{table}[h]
      \begin{minipage}[t]{0.45\textwidth}
        \centering
        \begin{tabular}{l|rr}
                         &Completed wafers & Tardiness \\ \hline
                    \ac{FIFO}  & 98.1  & 100.0     \\
                    \ac{CR}    & 99.5  & 68.2      \\
                    \textbf{\ac{SRPT}}  & \textbf{100.0} & \textbf{61.5}      \\
                    \ac{SPT}   & 98.8  & 87.1      \\
                    \ac{EDD}   & 99.2  & 73.5     
                \end{tabular}
        \caption{Performance of different heuristics for the Minifab model, normalized using the highest value for each metric.}
        \label{tab:heuristics_Minifab}
      \end{minipage}
      \hfill
      \begin{minipage}[t]{0.45\textwidth}
        \centering
        \begin{tabular}{l|rr}
                      & Completed wafers & Tardiness \\ \hline
                      \ac{FIFO} & \textbf{100.0} & 3.2\\
                      \textbf{\ac{CR}}   & 98.3  & \textbf{2.4}\\
                      \ac{SRPT} & 81.8  & 87.8\\
                      \ac{SPT}  & 79.1  & 100.0\\
                      \ac{EDD}  & 91.3  & 25.3\end{tabular}
        \caption{Performance of different heuristics for the SMT2020 model, normalized using the highest value for each metric.}
        \label{tab:heuristics_smt2020}
      \end{minipage}
    \end{table}
    The results for SMT2020 in Table~\ref{tab:heuristics_smt2020} yield that the
    \ac{CR} heuristic performs best regarding the tardiness of lots, which is our
    main metric to optimize.
    Hence, although \ac{FIFO} has a slight advantage in the number of completed wafers,
    we select \ac{CR} as the baseline dispatching heuristic for the SMT2020 fab models.
        
    \subsection{Experiments with \ac{PPO} Algorithm} \label{sec:ppo}
        
        Our first series of \ac{RL} experiments investigates the behavior during training of a dispatching strategy for the lithography area. All other tools are controlled by the default dispatching heuristics. It has to be noted that, while the number of overall tools and the number of lithography tools are somewhat comparable for SMT2020 and our industry scenario, it is just one tool in the case of the Minifab. However, as the model only has a total number of five tools, the impact of only controlling the lithography tool is expected to be very significant.
        
        The hyperparameters and the reward have to be adapted for the individual models. This is because the cause-effect relationship is delayed for the larger models as more decisions are made per day. Since the frequency of the KPI-based reward feedback per simulated day is constant but the number of actions changes, more actions are performed in-between reward signals. Secondly, the larger models have a more stable WIP level and the reward signal is less noisy. In order to account for this, the parameters are adapted as follows.
        The rollout fragment length, also called horizon, is adapted to be roughly the equivalent of 1-2 days of dispatching. This parameter defines the number of actions per worker and sample over which the advantage is calculated.
        
        \begin{figure}[t]
            \centering
            \includegraphics[width=0.99\textwidth]{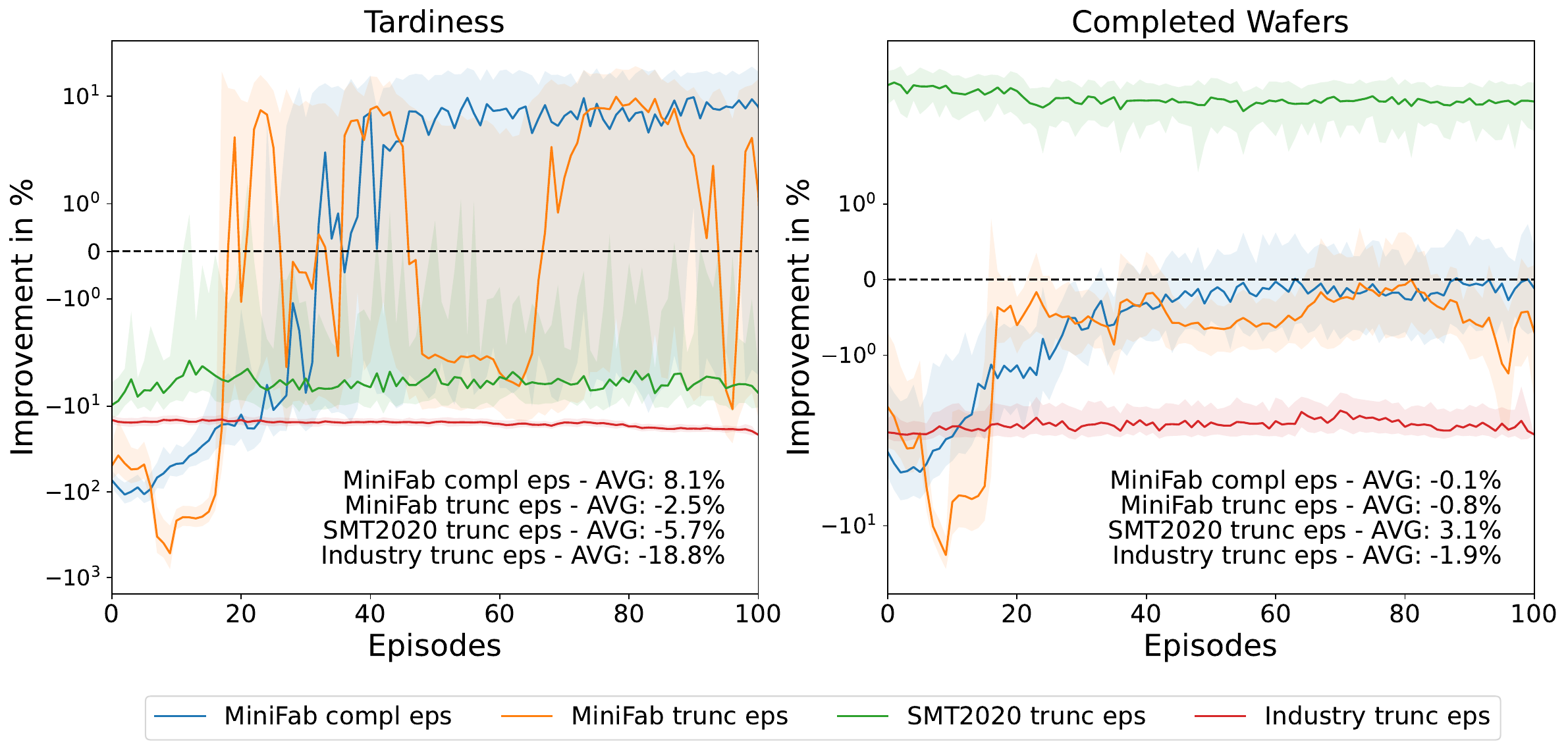}
            \caption{Results for the PPO experiments controlling the lithography area for three different simulation models with truncated and completed episodes for policy updates.}
            \label{fig:ppo_training}
        \end{figure}
        
        Another important hyperparameter is the truncation of episodes. If it is true (truc eps in Figure \ref{fig:ppo_training}), we perform multiple policy updates per episode, while not taking all samples collected from the entire episode into account. With our settings and setup, using complete episodes is only possible for the MiniFab, as the size of the collected samples becomes too big to handle for the larger models.
        Complete episodes are thus only considered for the Minifab model in Figure \ref{fig:ppo_training}, plotting the improvement in percent relative to the baseline dispatching heuristic for the tardiness and throughput metrics over the training progress in episodes. 
        Please note that the truncation of episodes does not change the simulation horizon but only the number of samples considered for a policy update of the PPO. While we show only the results for the best performing reward in Figure \ref{fig:ppo_training}, additional rewards are compared in the appendix under Section \ref{sec:ppo_appendix}.
        
        In one episode, we collect about 50 rollouts per worker. For the Minifab model, each rollout consists of 16 steps, where each step contains an observation for each lot in the queue. The queue length varies between one and a few hundred lots. For SMT2020, each rollout contains over 3000 steps, so that the observations become substantially more exhaustive. This means that the size of the collected samples is 100-1000 times higher, requiring much more memory and making the policy updates computationally expensive. However, using complete episodes leads to much more stable training progress for the MiniFab and could also be useful for the other models.
        \begin{wrapfigure}[18]{r}{0.6\textwidth}
           \begin{flushright}
            \includegraphics[width=0.43\textwidth]{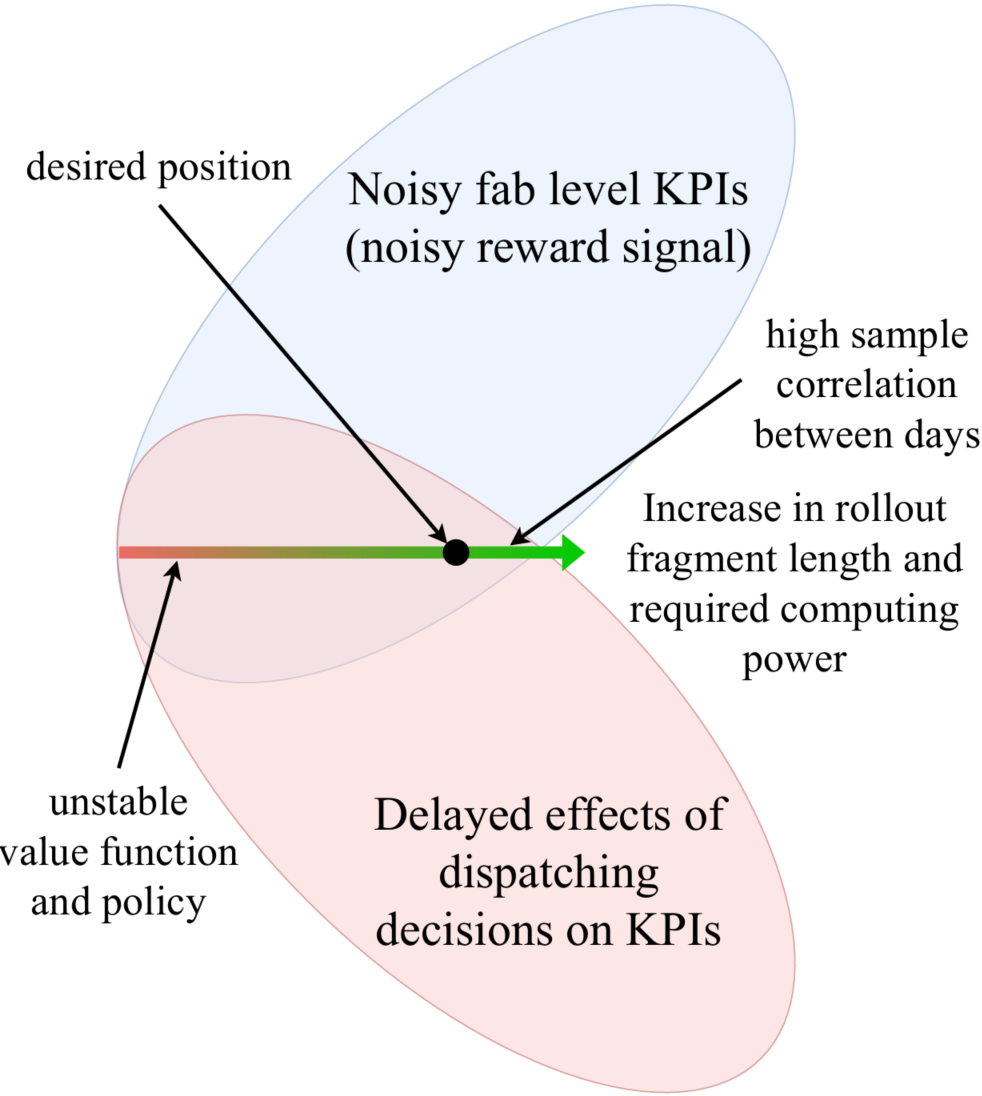}
            \captionsetup{justification=raggedleft,singlelinecheck=false}%
            \caption{Influences on PPO behavior.}
            \label{fig:stability_ppo}
           \end{flushright}
        \end{wrapfigure}
        It becomes clear from Figure \ref{fig:ppo_training} that we manage to achieve significant improvements for the Minifab model only. Reasons for this could be that the effects of individual actions are too delayed, the KPIs are too noisy, and the value function estimate is too inaccurate for the larger models. These issues are illustrated in Figure~\ref{fig:stability_ppo}.
        In case a long rollout fragment length is used to tackle the noisy reward signal and long-lasting effects of actions, the sample correlation among the steps collected by the same worker in the same rollout increases. This makes it harder for the algorithm to identify advantageous actions within the rollout. In turn, more parallel rollout workers would be needed to collect more samples, which leads to a drastic increase in required computing power and memory.

        While the realistic models have cycle times of many weeks and complex long-lasting effects, as well as an unstable situation due to ever-changing loading, the Minifab features constant loading and a short cycle time of only a few days. This explains why the Minifab achieves improvements under the settings of the experiments, while the other models fail. We suspect that, in order to make \ac{PPO} work for the larger models, the rollout fragment length as well as the number of workers have to be drastically increased, or we would need to find a smoother dense reward.
        
    \subsection{Experiments with \ac{ES} Algorithm}

\begin{figure}[t]
            \centering
            \includegraphics[width=0.99\textwidth]{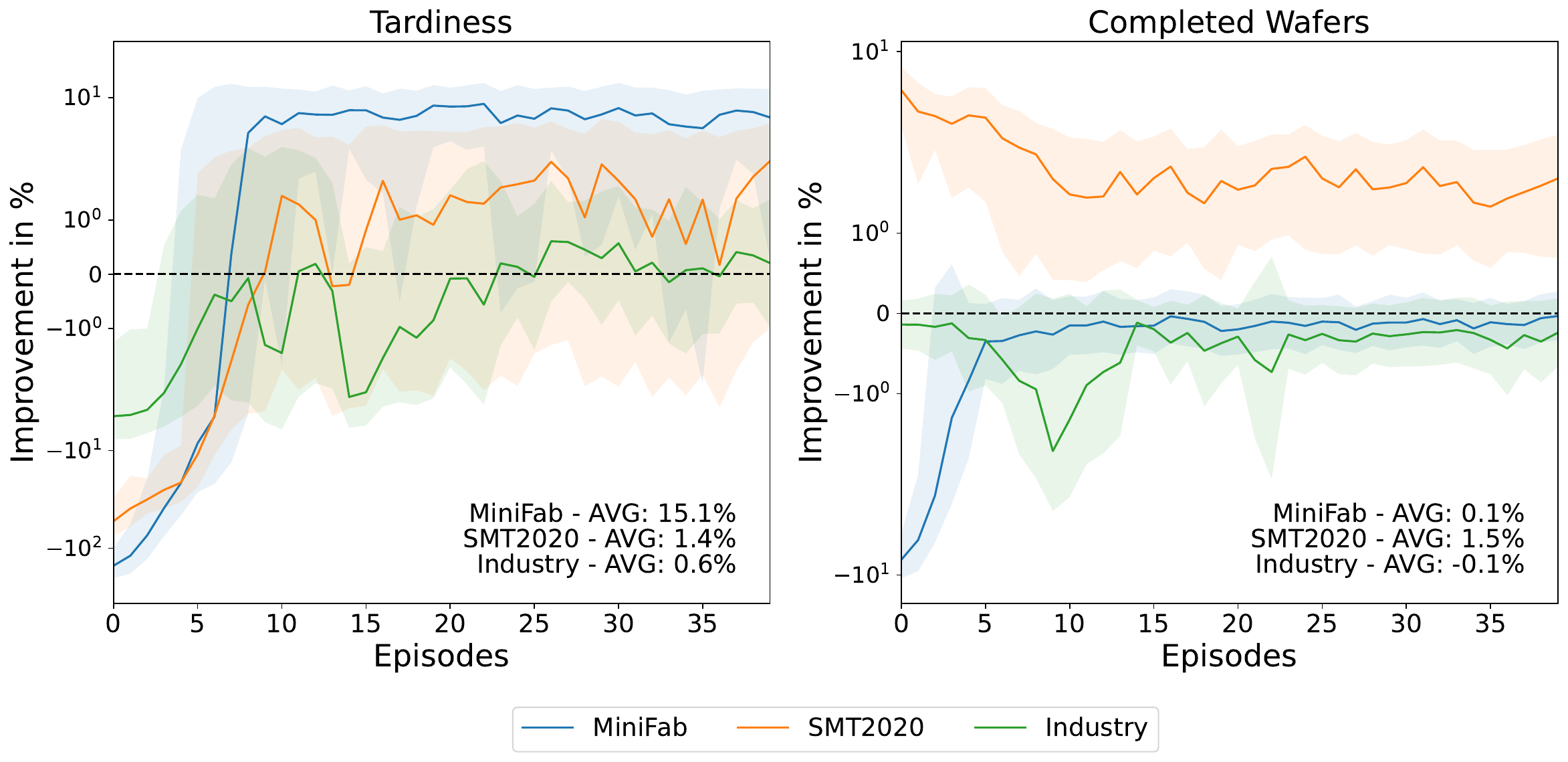}
            \caption{Results for the \ac{CMA}-\ac{ES} experiments controlling the lithography area for three different simulation models.}
            \label{fig:es_training}
        \end{figure}

        Figure~\ref{fig:es_training} follows the same notation as Figure~\ref{fig:ppo_training}, but instead of the \ac{PPO} optimizer, results for the \ac{CMA}-\ac{ES} training progress are shown. It can be seen that this strategy can handle the industry scenario and SMT2020 much better. Overall, this approach leads to better training results than \ac{PPO} but converges slower. 

        \begin{figure}[t]
            \centering
            \includegraphics[width=0.99\textwidth]{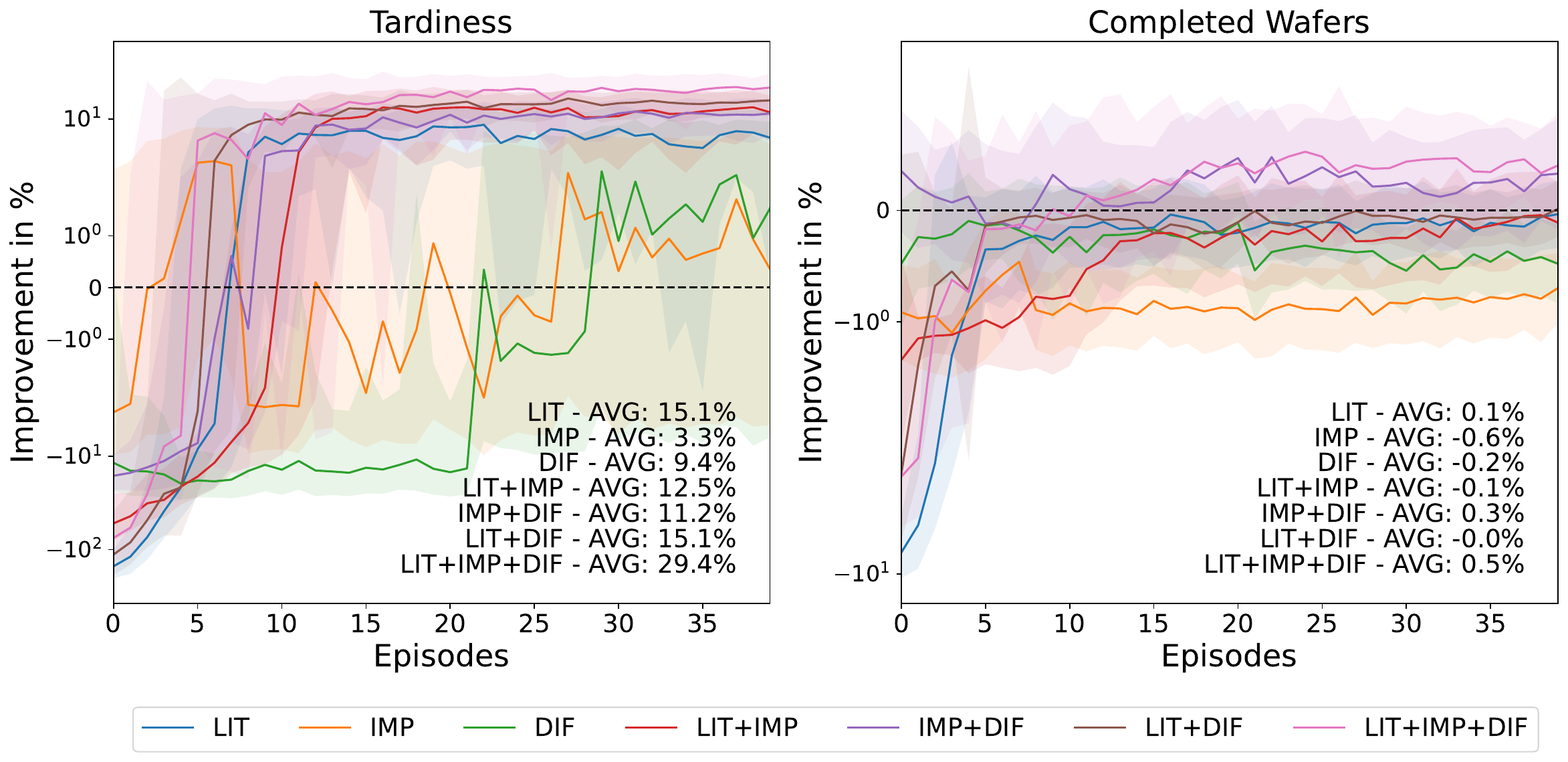}
            \caption{Results for the \ac{CMA}-\ac{ES} experiments controlling different combinations of tools for the Minifab model.}
            \label{fig:es_training_Minifab_all}
        \end{figure}

        Figure \ref{fig:es_training_Minifab_all} displays training results for the Minifab model with different combinations of controlled tools. It can be seen that the best results are achieved for the control of the entire system of three different tool types. It is further noticeable that optimization regarding the control of the diffusion tools and their batching process is only effective in combination with the other tools. This can be explained by the fact that the effective building of batches is dependent on the WIP supply from the previous steps. \\
        Similar behavior can be observed for SMT2020 and the industry scenario (see Figure~\ref{fig:es_training_smt2020_all} and Figure~\ref{fig:es_training_industry_all} in the appendix). That is, the higher the percentage of \ac{RL}-controlled tools, the higher is the potential improvement. The results are even better when training runs for 200 instead of 40 episodes only (see Figure~\ref{fig:es_training_Minifab_all_200} in the appendix), which comes as a matter of the computational cost.  

    \subsection{Computation Time}
\begin{figure}[t]
            \centering
            \includegraphics[width=0.99\textwidth]{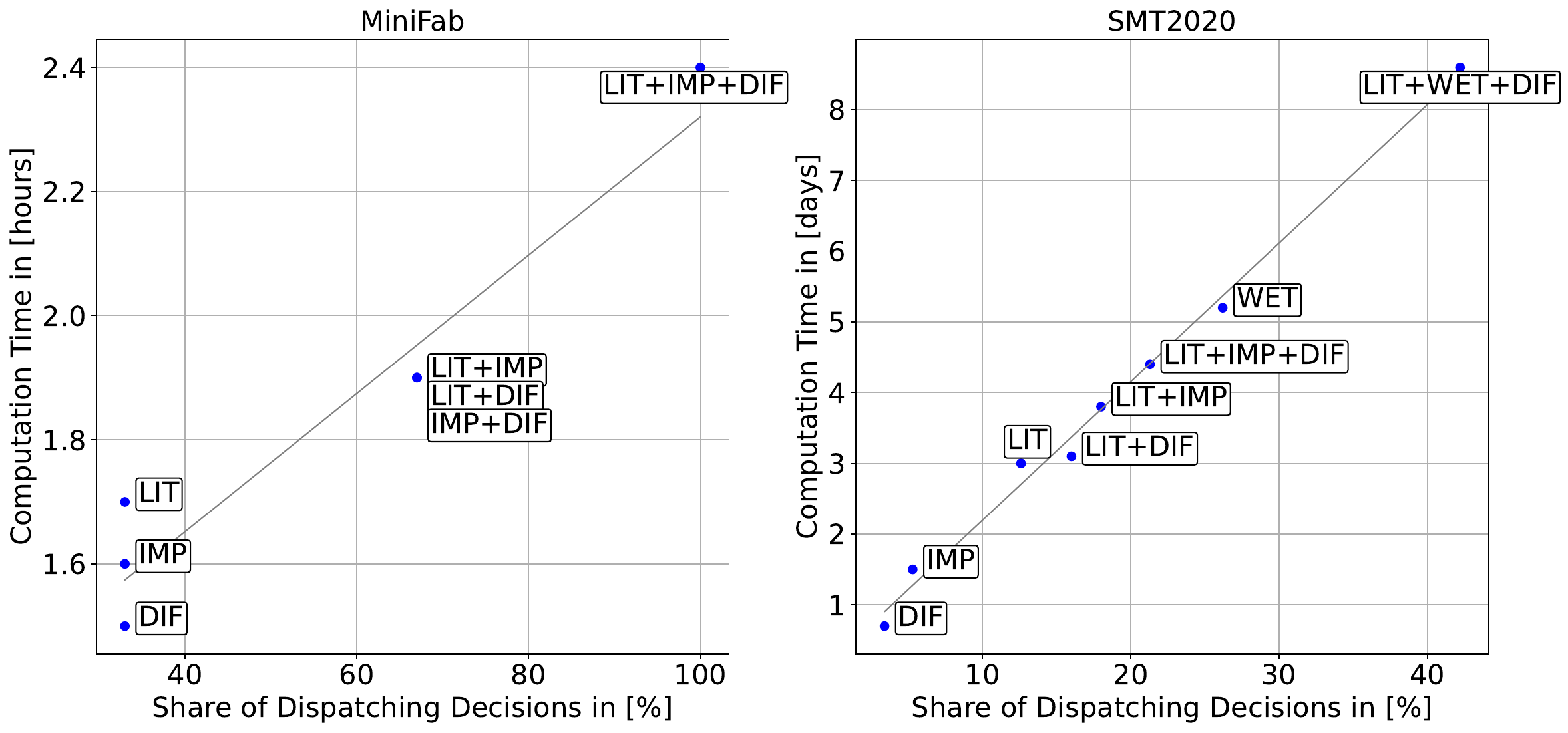}
            \caption{Computation times for the MiniFab and SMT2020 models for 40 training episodes.}%
            \label{fig:computational_cost}%
        \end{figure}

        Different sizes of the simulation models lead to vastly different execution and training times, which are plotted in Figure~\ref{fig:computational_cost}. Besides the size of the simulation model, the number of RL-controlled tools is relevant. Each of those tools has to communicate through the interface with the simulator for making dispatching decisions, which requires the extraction of an observation and the computation of a dispatching decision using the \ac{NN}.
        In fact, the plots show that scaling up the size leads to a growth of the execution time from hours to days.
        It has to be noted that the allocated memory scales linearly with the number of used CPU cores, as we assign one independent simulator instance to each core. Since the communication overhead increases with more parallel processes, this can lead to a longer execution time for individual episodes.
        The computational cost can be a factor in deciding on the utilized resources. However, for productive use in semiconductor frontend WIP flow management, the optimization potential is enormous. Fabs cost millions of dollars per week, and increasing the utilization of these resources by even one percent amounts to a significant financial advantage.
    
    \subsection{Generalization}
        In order to investigate the generalization capabilities of the different approaches, we first define the two main sources of uncertainty. The first one includes the stochastic effects of the daily operations, such as unplanned machine breakdowns. This is a huge factor in semiconductor manufacturing as machines fail frequently. It is handled by modeling the failure behavior of the tools with distributions and drawing random values from that distribution based on the random seed for each simulation run.

        \begin{figure}[b]
            \centering
            \includegraphics[width=0.99\textwidth]{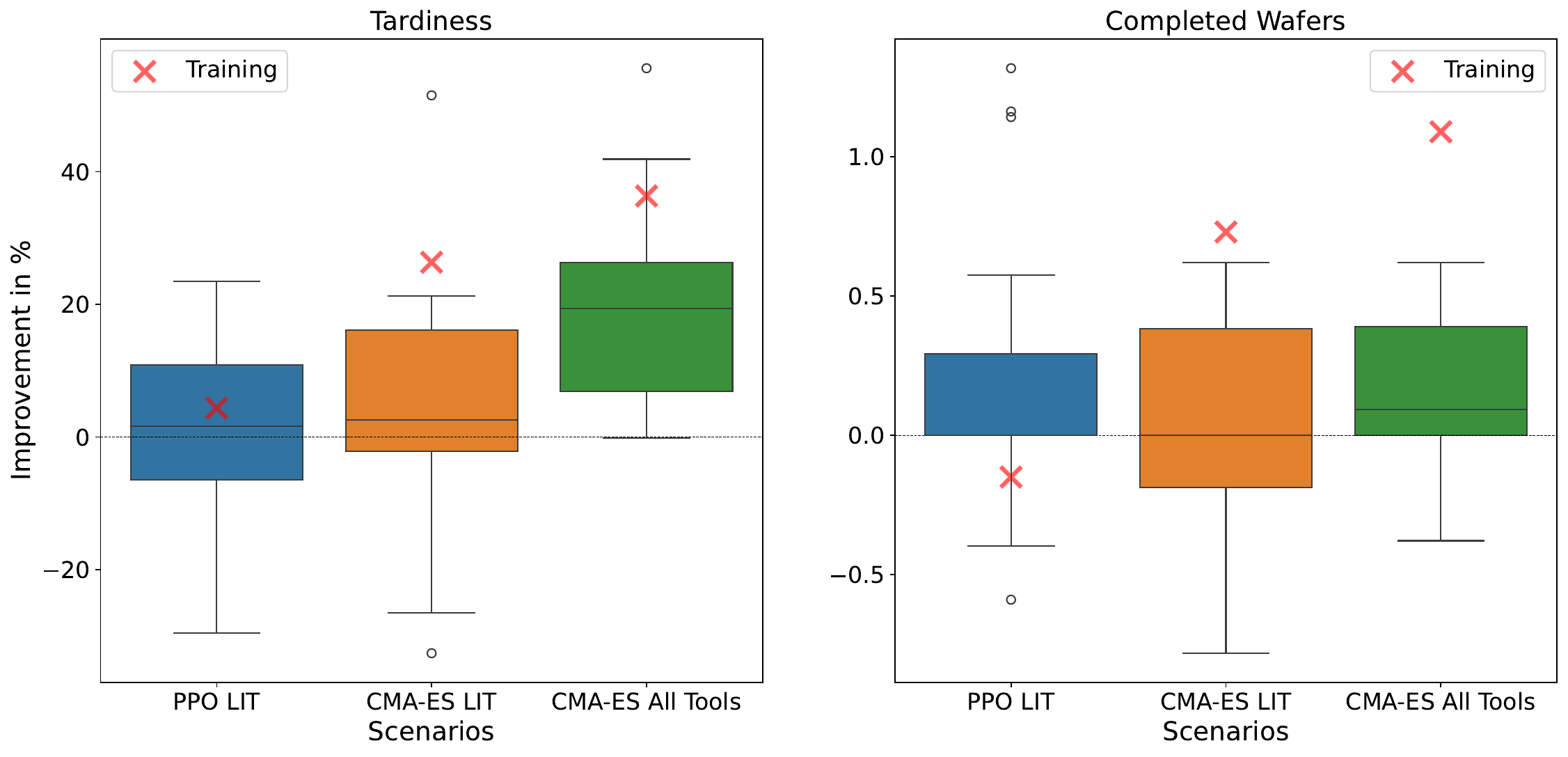}
            \caption{Results for the testing of the \ac{CMA}-\ac{ES} experiments after 200 episodes, controlling lithography tools or all tools for the Minifab, as well as the \ac{PPO} experiment controlling lithography tools. The red crosses indicate the result of the best parameter set of the last episode during the training for \ac{CMA}-\ac{ES}. For \ac{PPO}, the red crosses indicates the mean of the training results for the last episode as it is only executing one parameter set in parallel but on different random seeds for the simulation.}
            \label{fig:es_ppo_testing_Minifab}
        \end{figure}
        
        Figure \ref{fig:es_ppo_testing_Minifab} shows the generalization properties for the MiniFab model, which was trained with \ac{CMA}-\ac{ES} on one random seed for the lithography tool only, and once for all tools. It shows that the agent is able to generalize with some loss of improvement if it controls enough tools. Training with more random seeds does not help in this case, as our experiments exhibited a smaller gap between training and testing but much worse training results when training on more random seeds. This could be due to the noise that is injected by the use of more random seeds.

        \begin{figure}[t]
            \centering
            \includegraphics[width=0.99\textwidth]{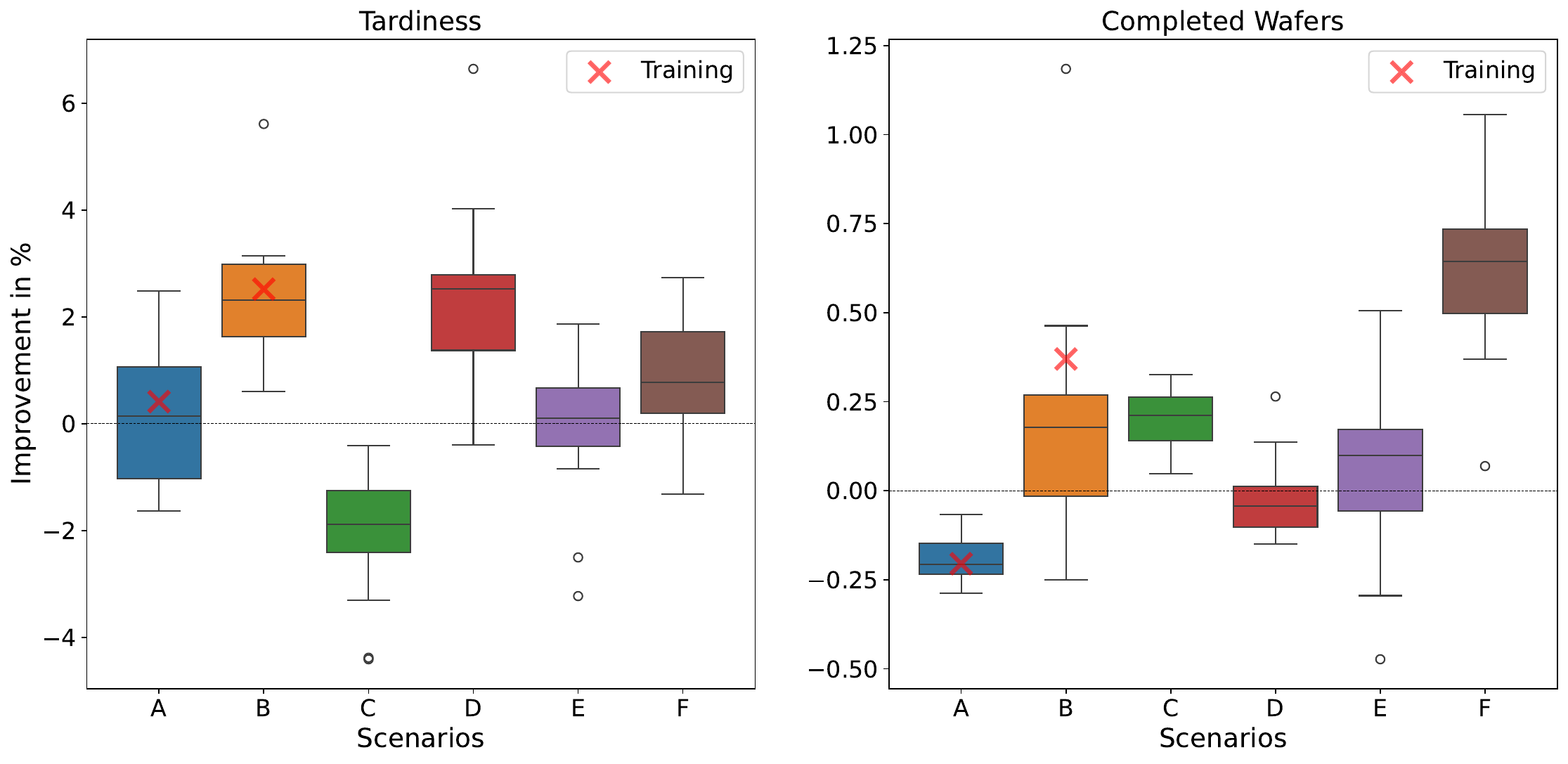}
            \caption{Results for the testing of the \ac{CMA}-\ac{ES} experiments controlling lithography and WET tools for the industry model. The red crosses indicate the result of the best parameter set of the last episode during the training. The scenarios without red crosses were not included in the training.
            }
            \label{fig:generalization_cma_industry}
        \end{figure}
        
        For \ac{PPO}, we already train on many different random seeds as we do not have to use our cores to test multiple perturbations of the \ac{NN} parameters in parallel. For this reason, the strategy generalizes quite well on the test runs. 
        If we train on only one random seed, the \ac{PPO} delivers much worse performance. This could be because the noise of multiple seeds helps to not overfit on the reward of individual time intervals. However, the training and testing result is worse than for the \ac{CMA}-\ac{ES} approach. While more random seeds cannot be handled by the ES algorithm, \ac{PPO} has no problems with the induced noise. This could be due to the fact that the \ac{PPO} algorithm is not a black-box algorithm and utilizes the information collected from each decision step, not only from the KPIs achieved over the entire episode.
        
        The second source of uncertainty is the ever-changing load mix. The volume and mix of products are changing over time and can only be estimated for the near future. This issue is only present in the SMT2020 and industry models, while the MiniFab model uses a cyclic pattern for the lot release.

        In order to test how strategies found by the agent are generalizing for different load mixes, we test the \ac{CMA}-\ac{ES} strategy on multiple different loading scenarios. However, one could also decide to retrain a specific strategy for each new loading scenario in regular planning cycles. This would possibly lead to less robust but ideally more optimized strategies for the individual cases.

        The test results for training the \ac{CMA}-\ac{ES} strategy with two scenarios and two random seeds each are shown in Figure~\ref{fig:generalization_cma_industry}, where the agent is controlling the lithography and WET areas.
        These results again indicate that the agent is generalizing with some loss of improvement. It can also be noted that the agent over-optimizes one training scenario more than the other. Furthermore, at least one scenario is always a bit worse than the reference for each metric. We suspect that these phenomena can be mitigated by training on more random seeds and scenarios in parallel.

       SMT2020 does not provide different loading scenarios for the same production system under the same conditions. The models differ in the number of tools and the types of considered lots. Therefore we only test the trained model on the same scenario but on a longer time horizon (6 months instead of 50 days). This is useful as the loading changes over time.

       \begin{figure}[t]
            \centering
            \includegraphics[width=0.99\textwidth]{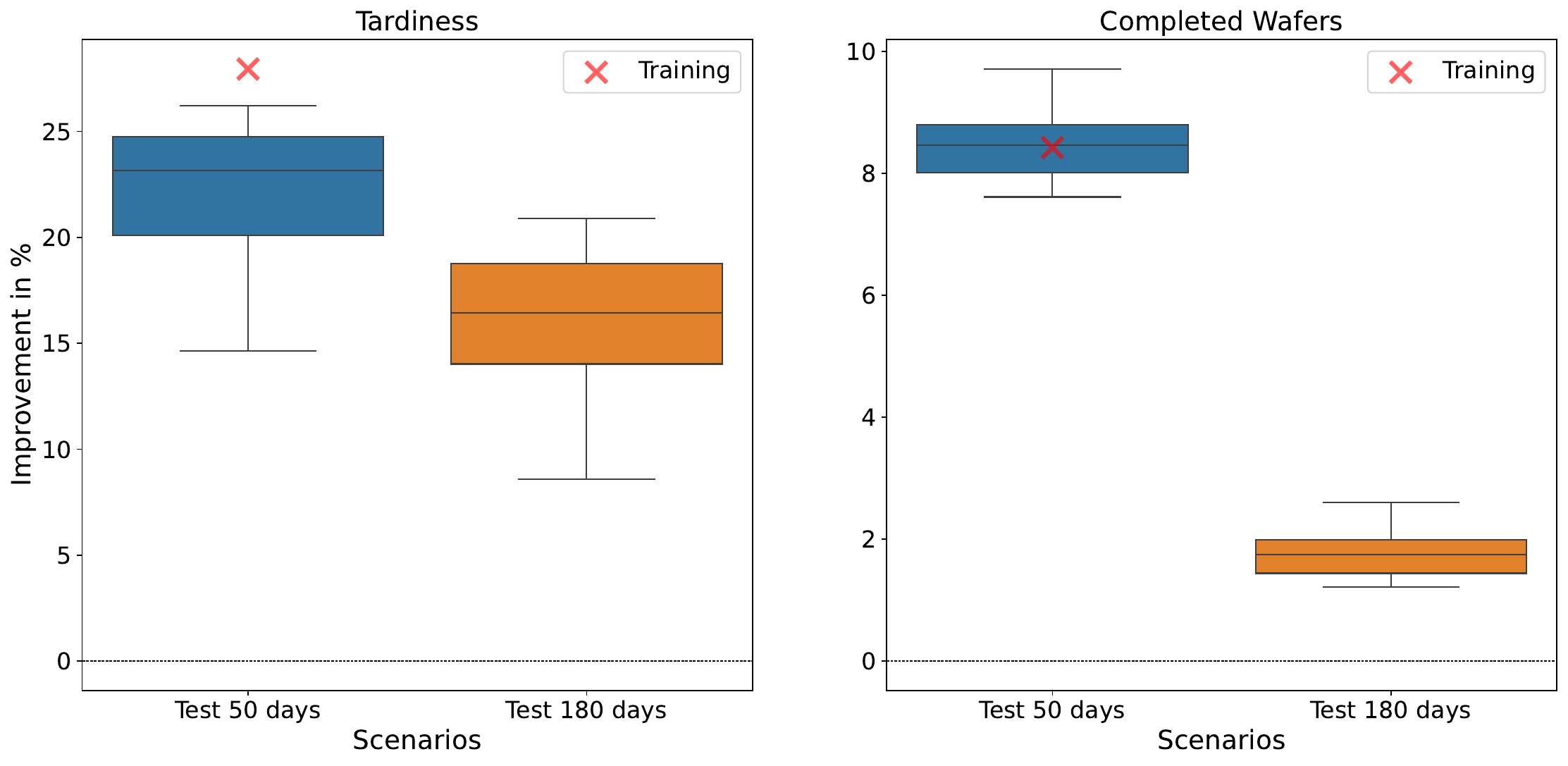}
            \caption{Results for the testing of the \ac{CMA}-\ac{ES} experiments controlling all tools for the SMT2020 model. The red crosses indicate the result of the best parameter set of the last episode during the training, where the agent is trained on the 50-day scenario only.}
            \label{fig:generalization_cma_smt2020}
        \end{figure}

        In Figure~\ref{fig:generalization_cma_smt2020}, we observe that the \ac{CMA}-\ac{ES} strategy yields improvements with respect to the reference for both metrics, yet with some loss for the longer unseen scenario. This gap could be closed by training directly on the longer scenario, which would be computationally more expensive as each episode takes roughly four times longer to simulate. It has to be noted that, similar to the results for the MiniFab in Figure \ref{fig:es_ppo_testing_Minifab}, SMT2020 also does not show good generalization properties if only one tool type is controlled by the agent.

\section{Conclusion}\label{sec:conclusion}

    While the amount of publications on \ac{RL} in the context of semiconductor manufacturing is increasing every year, they substantially lack comparability. Different authors work on different model scenarios as well as different simulators, and utilize a variety of algorithms on diverse hardware. Thus it is hard to assess the scalability of an \ac{RL} algorithm for dispatching found in the literature. In order to mitigate this shortcoming, we tested the scalability of an \ac{ES} approach and a classical policy-gradient approach to optimize dispatching. We compared the performance of these algorithms across two public benchmark datasets of significantly different sizes and on an industry dataset. 
    
    The most important findings are the limited scalability of the \ac{PPO} approach for this problem setting, which only shows an improvement for the Minifab model. \ac{ES} scales well with an increased number of CPU cores. Furthermore, it is significantly harder to optimize real-world scenarios and the potential improvement is smaller. When switching between models, the algorithm has to be tuned regarding the hyperparameters and the reward function. Lastly, the higher the percentage of tools under the control of the \ac{RL} system, the higher is also the potential improvement. For the two open-source benchmark models Minifab and SMT2020, we achieve double-digit percentage improvement in tardiness and single digit percentage improvement in throughput. For the real industrial scale scenarios, we achieve an improvement of up to 4\,\% regarding tardiness and up to 1\,\% regarding throughput.

\section{Limitations and Future Research}
    The use of policy updates with state-action-reward samples from complete episodes is shown to be more stable and yields better results than truncated samples in our study. However, it was computationally not possible with our setup to save all samples from an entire episode for the industrial scale scenario due to memory overflow. This is due to millions of samples which each have a big observation space. However, with more dedicated hardware resources, the PPO potentially would deliver more promising result for the industrial scale scenario. With more computing power and memory, we could also extend the experiments for the industrial scale model to utilize the RL-agent for all dispatching decisions, not only for bottleneck equipment groups.  As different tools might require different dispatching strategies and the tool types are too many to efficiently encode in the observation space, the approach might have to be adapted to a multi-agent system with one policy per tool type. 
    
    As we have shown that CMA-ES is an alternative to policy gradient methods in \ac{SFSP} and scales better with the complexity of the simulation model, the findings of this work can help to scale the training more effectively and thus reduce the overall computational cost. 
    
    Potential future areas of research include Explainable AI (XAI) to communicate the results to stakeholders like the responsible managers of the production facilities. Furthermore, there is a need for well-planned testing and validation steps to ensure the safety and generalization in the real production system outside of the simulation environment.
    
\section*{Statements \& Declarations}
    
    \bmhead{Data availability statement} 
        Due to commercial reasons, the real industry dataset is not available. However, the original dataset of the SMT2020 model \citep{Kopp.2020} and descriptions for the MiniFab model \citep{ElKhouly.2009, Spier.1995} are already available.
    
    \bmhead{Funding}
        This work was partially funded by the German Federal Ministry for Economic Affairs and Climate Action (BMWK) project 13IK033 (SmartMan) as well as the Austrian Research Promotion Agency (FFG) projects 894072 (SwarmIn) and FO999910235 (SAELING).
        
    \bmhead{Competing Interests} The authors declare that they have no conflict of interest.
    
    \bmhead{Author Contributions} P.S. developed the initial concept and wrote the main manuscript. 
    P.S. and A.I. developed and implemented the method. This includes among others the neural network architecture, training procedure and environment, and the adaptation of the CMA-ES method to the use case. H.S. carried out the experiments, analyzed the results, validated and revised the implementations of the benchmark datasets. T.A. supervised the work from Infineon side and provided guidance regarding the manufacturing domain and methodology. M.W. contributed to the design of experiment for the comparison of the benchmark models and methodology for the literature search. M.G. and K.S. supervised the work from the academic side and provided theoretical guidance. G.S. provided guidance regarding the simulation and domain knowledge. C.W.C. implemented the benchmark datasets for the used simulator engine. C.W.C. and F.F.Z. provided support with the simulator and the simulator interface. All authors discussed the results and contributed to the final manuscript.

\backmatter

\begin{appendices}

\section{Comparison of Training Results for PPO Rewards}\label{sec:ppo_appendix}
\begin{figure}[H]
        \centering
        \includegraphics[width=0.95\textwidth]{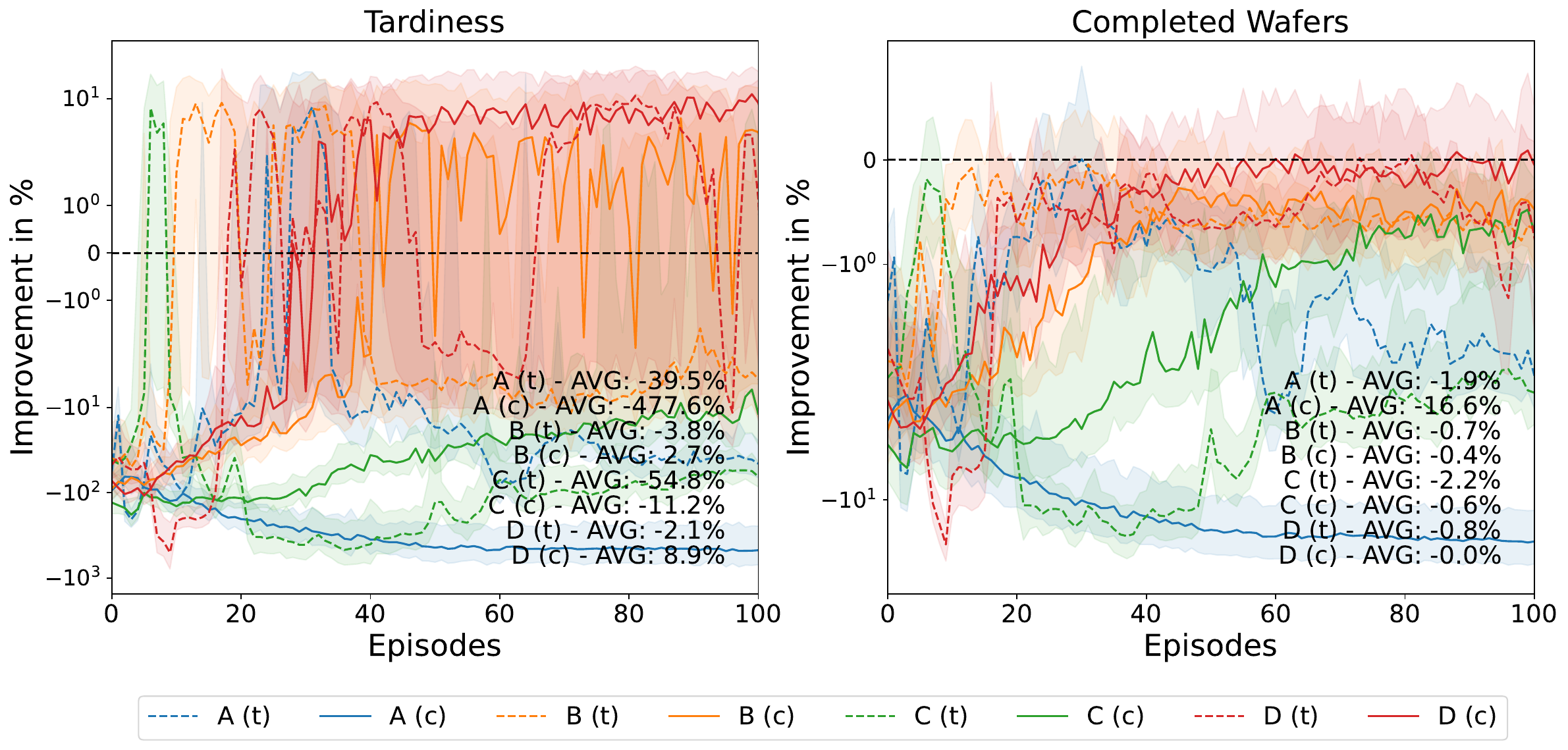}
        \caption{Results for the \ac{PPO} experiments with different rewards controlling the lithography area of the MiniFab model. The rewards are calculated hourly as a rolling mean of the previous 24\,h. We differentiate between completed (c) and truncated (t) episodes used for the PPO policy updates. The truncation does not affect the simulated horizon but only the number of samples which is stored and used for the back-propagation.}
        \label{fig:rewards_minifab}
    \end{figure}

\begin{figure}[H]
        \centering
        \includegraphics[width=0.95\textwidth]{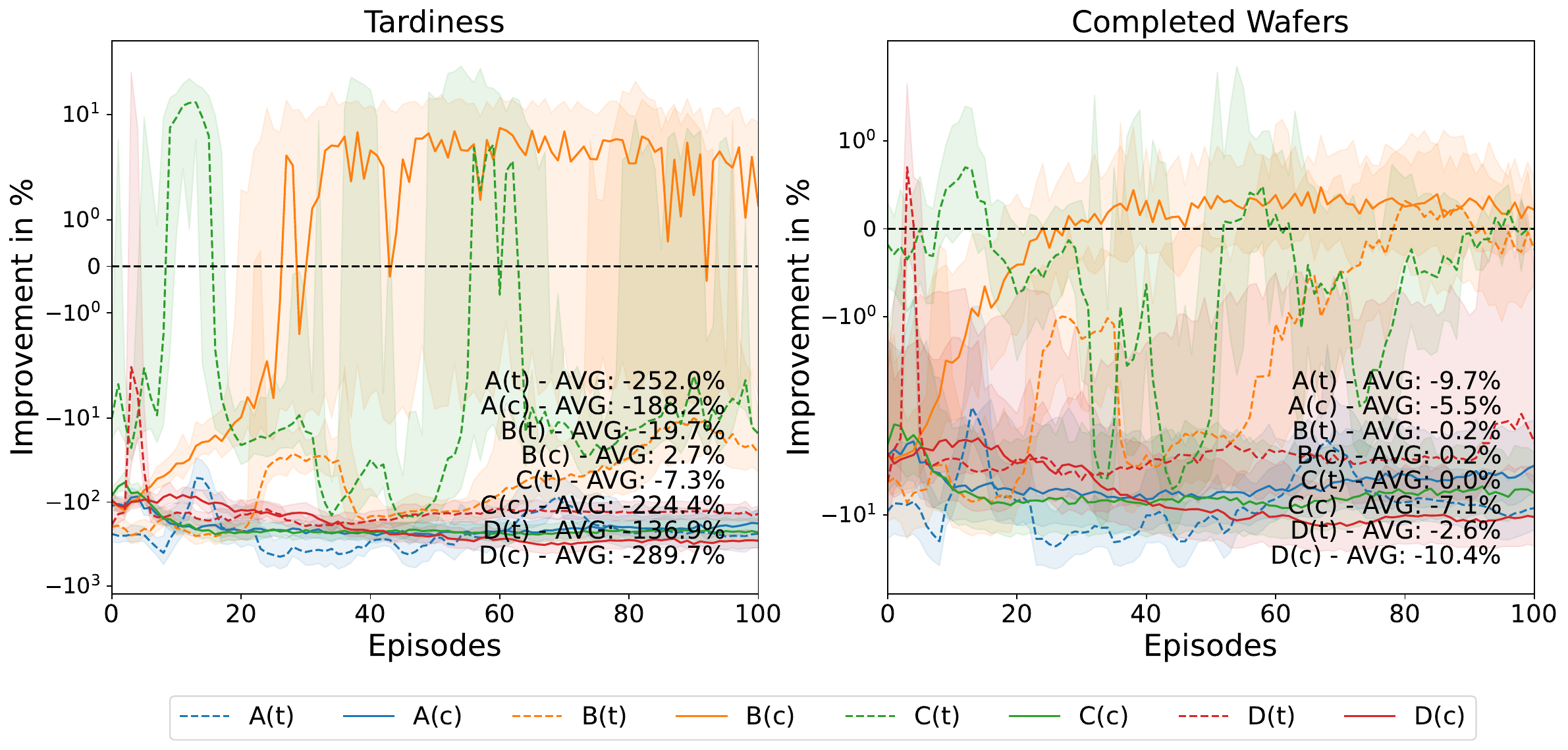}
        \caption{Results for the \ac{PPO} experiments with different rewards controlling the lithography area of the MiniFab model. The rewards are calculated hourly considering only the previous 60\,min. We differentiate between completed (c) and truncated (t) episodes used for the PPO policy updates. The truncation does not affect the simulated horizon but only the number of samples which is stored and used for the back-propagation.}
        \label{fig:rewards_minifab_hourly}
    \end{figure}

    We show the results for the best performing PPO reward $r_{\mathrm{PPO},t} = r_{\mathrm{D},t}$ in Section \ref{sec:ppo}. In the following, we introduce additional rewards which we initially compared to find the best performing one. The rewards are based on different combinations of the tardiness and throughput of the overall fab, with $tp_t$ being the number of completed wafers during the past 24 hours and $\mathrm{WIP}_t$ the number of wafers remaining in the system at time $t$, the end of the 24 hours interval. The tardiness is described by the tardiness $td_{out,t}$ of the completed wafers within the past 24\,h and the tardiness $td_{in,t}$ of the wafers remaining in the system at at time $t$. The results of the comparison are shown in Figure \ref{fig:rewards_minifab}. The rewards are defined as follows:

    \begin{equation*}
        r_{\mathrm{A},t} = tp_t 
    \end{equation*}

    \begin{equation*}
        r_{\mathrm{B},t} = td_{in,t}
    \end{equation*}

    \begin{equation*}
        r_{\mathrm{C},t} = \frac{tp_t \cdot (\mathrm{WIP}_t + tp_t)}{tp_t \cdot td_{out,t} + \mathrm{WIP}_t \cdot td_{in,t}}
    \end{equation*}

    \begin{equation*}
        r_{\mathrm{D},t} = \frac{tp_t}{(tp_t \cdot td_{out,t} + \mathrm{WIP}_t \cdot td_{in,t})\cdot (\mathrm{WIP}_t + tp_t)}
    \end{equation*}
    In order to validate the approach using the rolling mean, we generate the results shown in Figure \ref{fig:rewards_minifab_hourly} as a comparison, for which we calculate the rewards not hourly over the past 24\,hours but only over the past hour. It can be seen that the training is much more unstable as the reward is more noisy. However, for reward $r_{\mathrm{B},t}$, it converges even faster. We suspect that this is due to the fact that $td_{in}$ is a stable metric because it is calculated over the entire WIP, not only over the completed wafers. Especially for the rewards which are a combinations of multiple KPIs, the noisy feedback is a problem if the denominator approaches 0 in individual time intervals.

\section{Additional \ac{CMA}-\ac{ES} Training Results}\label{secA1}

    \begin{figure}[H]
        \centering
        \includegraphics[width=0.95\textwidth]{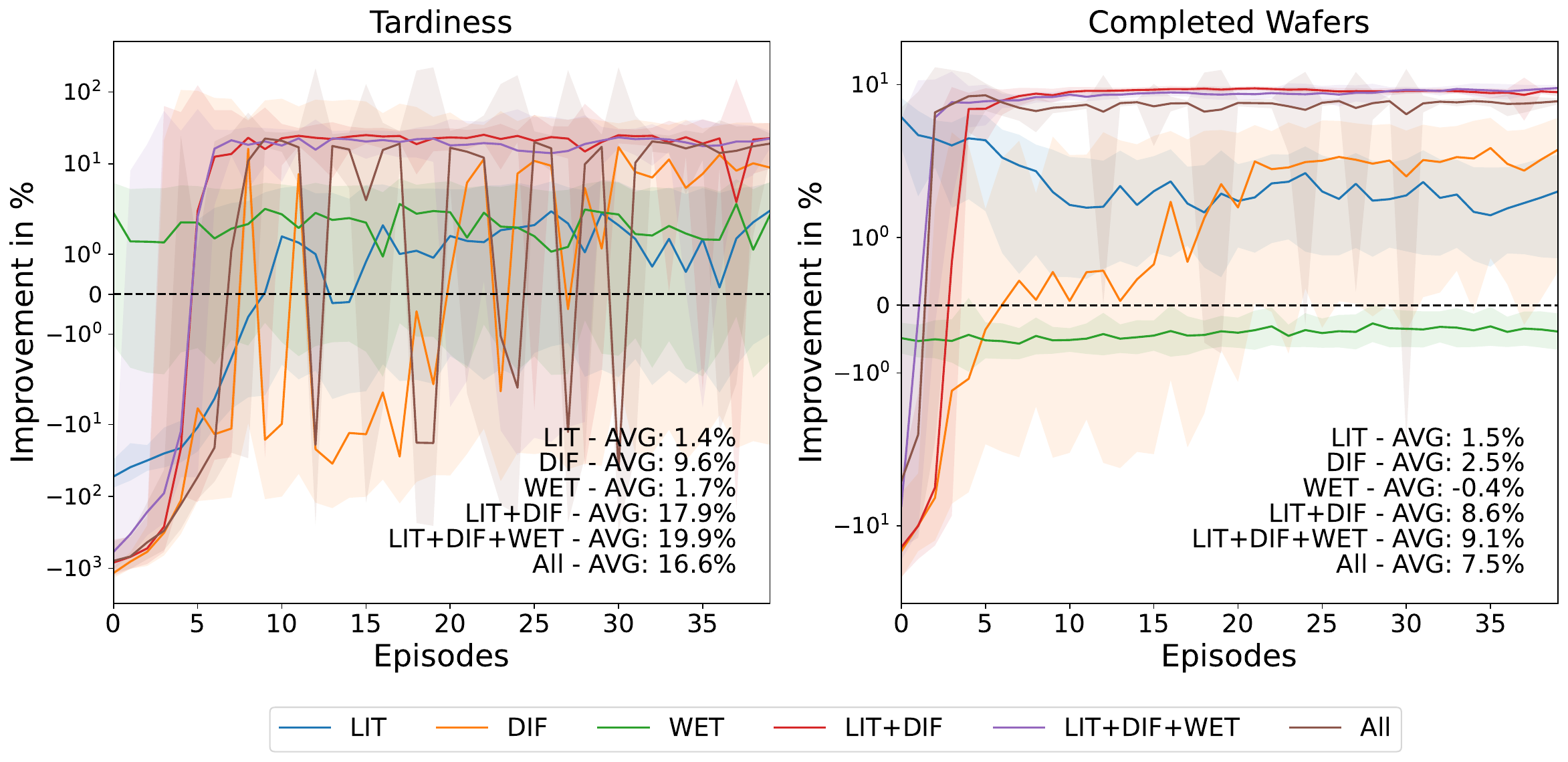}
        \caption{Results for the \ac{CMA}-\ac{ES} experiments controlling different combinations of tools for the SMT2020 model.}
        \label{fig:es_training_smt2020_all}
    \end{figure}

    \begin{figure}[H]
        \centering
        \includegraphics[width=0.95\textwidth]{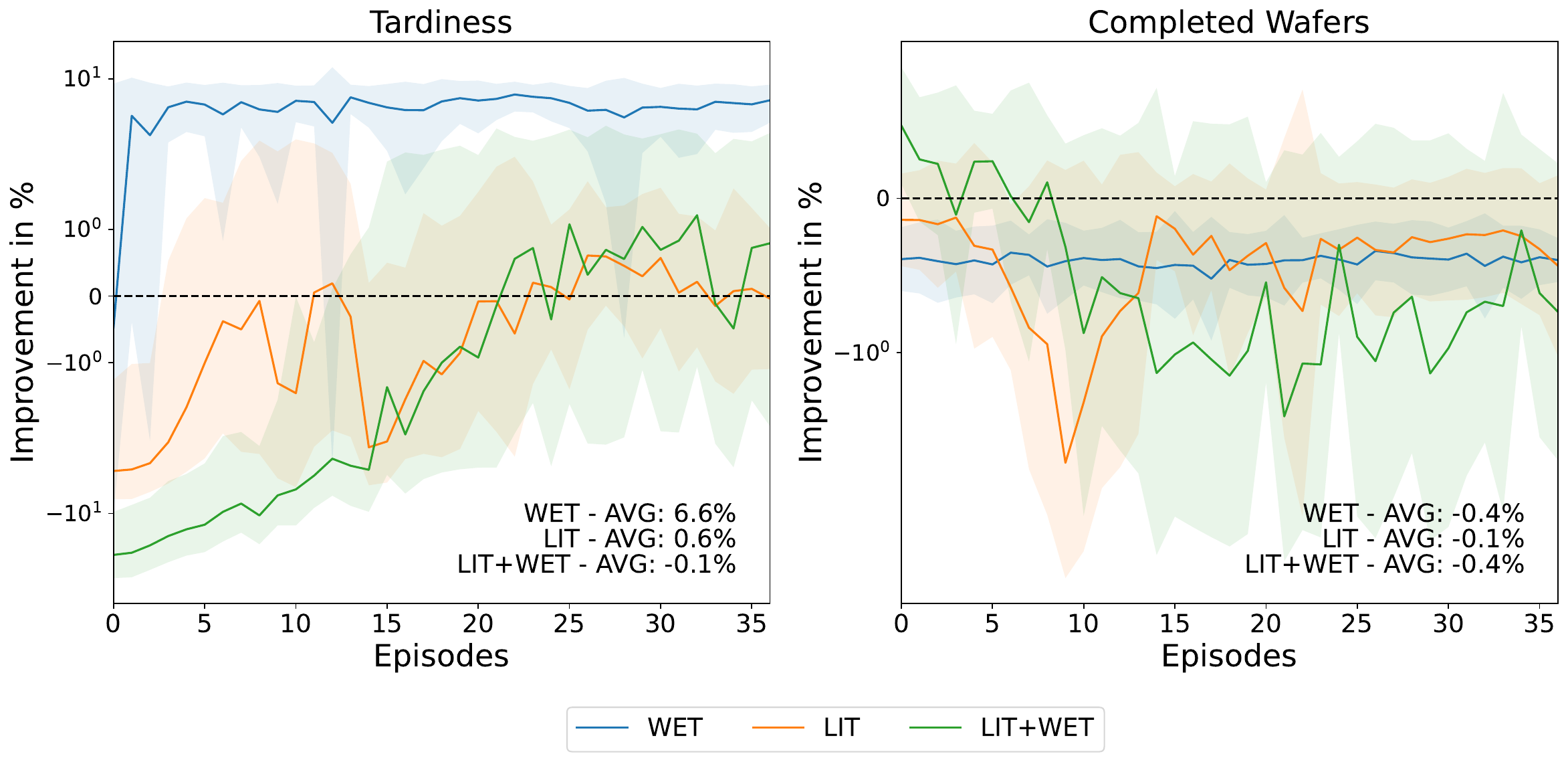}
        \caption{Results for the \ac{CMA}-\ac{ES} experiments controlling different combinations of tools for the industry model.}
        \label{fig:es_training_industry_all}
    \end{figure}

    \begin{figure}[H]
        \centering
        \includegraphics[width=0.95\textwidth]{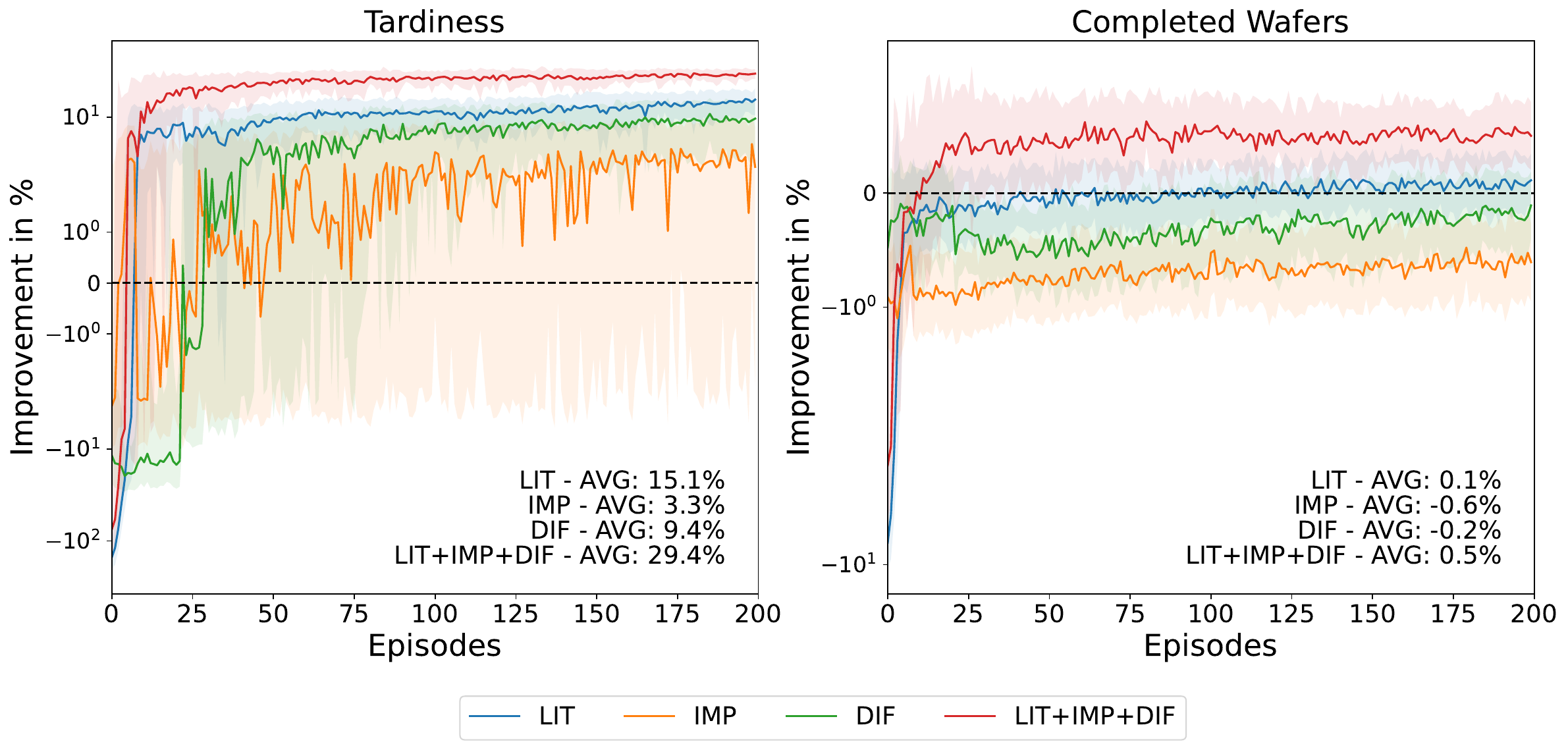}
        \caption{Results for the \ac{CMA}-\ac{ES} experiments after 200 episodes, controlling different combinations of tools for the Minifab model.}
        \label{fig:es_training_Minifab_all_200}
    \end{figure}

\end{appendices}

\bibliography{sn-article}

\pagebreak

\end{document}